\documentclass[review]{elsarticle}

\usepackage{lineno,hyperref}

\journal{Journal of \LaTeX\ Templates}

\usepackage{times}
\usepackage{helvet}
\usepackage{courier}
\usepackage{url}
\usepackage{subfigure}
\usepackage{graphicx}
\usepackage{amsmath}
\usepackage{amssymb}
\usepackage{multirow}
\usepackage{threeparttable}
\usepackage{float}
\graphicspath{{figs/}} 

\begin{document}

\begin{frontmatter}

\title{BTS-DSN: Deeply Supervised Neural Network with Short Connections for Retinal Vessel Segmentation}

\author[mymainaddress]{Song Guo}
\author[mymainaddress,mysecaddress]{Kai Wang}

\author[mymainaddress,mythirdaddress]{Hong Kang}
\author[myfourthaddress]{Yujun Zhang}

\author[mymainaddress]{Yingqi Gao}\author[mymainaddress]{Tao Li\corref{mycorrespondingauthor}}
\cortext[mycorrespondingauthor]{Corresponding author}
\ead{litao@nankai.edu.cn}

\address[mymainaddress]{Nankai University, Tianjin, China}
\address[mysecaddress]{Key Laboratory for Medical Data Analysis and Statistical Research of Tianjin (KLMDASR)}
\address[mythirdaddress]{Beijing Shanggong Medical Technology Co. Ltd}
\address[myfourthaddress]{Institute of Computing Technology, Chinese Academy}

\begin{abstract}
Background and Objective: The condition of vessel of the human eye is an important factor for the diagnosis of ophthalmological diseases. Vessel segmentation in fundus images is a challenging task due to complex vessel structure, the presence of similar structures such as microaneurysms and hemorrhages, micro-vessel with only one to several pixels wide, and requirements for finer results.
\\
Methods:In this paper, we present a multi-scale deeply supervised network with short connections (BTS-DSN) for vessel segmentation. We used short connections to transfer semantic information between side-output layers. Bottom-top short connections pass low level semantic information to high level for refining results in high-level side-outputs, and top-bottom short connection passes much structural information to low level for reducing noises in low-level side-outputs. In addition, we employ cross-training to show that our model is suitable for real world fundus images.
\\
Results: The proposed BTS-DSN has been verified on DRIVE, STARE and CHASE\_DB1 datasets, and showed competitive performance over other state-of-the-art methods. Specially, with patch level input, the network achieved 0.7891/0.8212 sensitivity, 0.9804/0.9843 specificity, 0.9806/0.9859 AUC, and 0.8249/0.8421 F1-score on DRIVE and STARE, respectively. Moreover, our model behaves better than other methods in cross-training experiments.
\\
Conclusions: BTS-DSN achieves competitive performance in vessel segmentation task on three public datasets. It is suitable for vessel segmentation. The source code of our method is available at \url{https://github.com/guomugong/BTS-DSN}.
\end{abstract}


\end{frontmatter}

\section{Introduction}
Ophthalmologic diseases, including age-related fovea degeneration, diabetic retinopathy, glaucoma, hypertension, arteriosclerosis and choroidal neovascularization, are the main causes for blindness. Thus, early diagnosis and treatment of these diseases are critical for reducing the risk of blindness \cite{Karpecki2015Kanski}. As an important component in retinal fundus images, vessel condition is always used as an indicator by ophthalmologists for diagnosing various ophthalmologic diseases. Since manual annotation of blood vessels in retinal images is tedious and time-consuming and requires experience in clinical practice, automated vessel segmentation is urgent \cite{Kirbas2004A}.

Automated segmentation of retinal blood vessels has attracted significant attention over recent decades \cite{Fraz2012Blood}. However, vessel segmentation is a challenging task for computers due to the following reasons \cite{Fraz2012An}. 1) The structure of retinal vessels is complex, and there are huge differences between vessels in different local areas in terms of size, shape and intensity. The first challenge is to build a model that can describe the complex vessel structure. 2) Some structures such as striped hemorrhage have similar shape and intensity with vessels. Furthermore, a micro-vessel is thin, whose width usually ranges from one to several pixels, making it easy to be confused with noises. The second challenge is to distinguish vessels from other similar structures or noises. 3) Compared to other object segmentation tasks such as Pascal VOC \cite{Everingham15}, vessel segmentation requires finer results to keep the structure of different vessels for subsequent vessel based diagnosis. The third challenge is to refine vessel pixels so that the structure of vessels can be well kept.

Existing segmentation algorithms could be divided into two categories \cite{Mo2017Multi}: unsupervised methods and supervised ones.

1) \emph{Unsupervised Methods}
Both features and rules have been designed manually by observing given samples in the unsupervised methods. For example, Azzopardi and Petkov
(2013) designed a 2D kernel used to convolve with retinal images by exploiting three characteristics of retinal blood vessels, including a piecewise linear approximation, a decrease in vessel diameter along vascular length, and a Gaussian-like intensity profile \cite{Azzopardi2013Automatic}. Yin, Adel, and Bourennane (2012) presented a vessel tracking algorithm by utilizing the continuity of vessels and local information to segment a vessel between two seed points, and the center of a longitudinal cross-section of a vessel is determined by gray-level intensity and tortuosity \cite{Yin2012Retinal}. Similarly, other characteristics have been successively exploited for vessel segmentation, such as profile \cite{Wang2007Analysis}, contour \cite{Al2009An} and geometric \cite{Sum2008Vessel}. The advantage of these unsupervised methods is low labor cost since label information is not required for the given samples. However, the unsupervised methods are sensitive to the manually designed features and rules. When the given sample set is becoming larger, there is an increasing difficulty for human to design the best features and rules for all given samples. Therefore, the unsupervised methods may show amazing performance in some samples but also poor performance in other samples and it is possible that a large sample set makes human confused on the design of features and rules.

2) \emph{Supervised Methods}
Vessel segmentation can be regarded as a pixel-wise binary classification problem, in which each pixel is classified as a vessel pixel or non-vessel pixel. Recently, deep learning is popular for pixel-wise classification tasks \cite{Melinsca2015Retinal,Khalaf2016Convolutional,Liskowski2016Segmenting}  due to its good performance in practical applications. However, since a retinal image always contains a large number of pixels ($>$100k) to be classified, such pixel-wise methods are time-consuming and difficult to satisfy real-time requirements of clinical diagnosis. Other methods are based on supervised semantic segmentation, capable of generating segmentation results by only one forward pass \cite{Mo2017Multi}. Fu et al. (2016) presented a method to segment vessels by combining HED (Holistically-nested Edge Detection) \cite{Xie2015Holistically} and conditional random field \cite{Fu2016DeepVessel}, in which HED is a popular FCN-based (Fully Convolutional Network) \cite{Long2017Fully} edge detection network. Maninis et al. (2016) designed a novel multi-task FCN network for simultaneous segmentation of vessels and optic disc \cite{driu}. However, due to the limit of memory, each image has to be down-sampled to a low resolution before it is fed into the model, which causes a coarse segmentation result. To obtain a fine result, patch-level methods are proposed  \cite{Li2016A,Oliveira2017Augmenting,feng2017patch,zhang2018,wu2018,jbhi2018,OLIVEIRA2018229}, in which each image is partitioned into multiple patches and each patch is fed into the model. After segmentation results for each patch are generated, they are puzzled into the final segmentation result. Patch-level segmentation is more time-consuming than image-level segmentation \cite{zhang2018}, but able to achieve higher performance due to a higher resolution. These supervised methods can automatically learn features or rules from a training set, and their performance is expected to be improved when the number of available training images is increased.

Supervised methods have achieved state-of-the-art results on vessel segmentation in previous works. However, deep supervision information is not well exploited to improve segmentation results although it has been widely used for other segmentation tasks. In this paper, we propose a deeply-supervised fully convolutional neural network with bottom-top and top-bottom short connections (BTS-DSN) that fuses multi-level features together to obtain fine segmentation results. Similar to other supervised methods, the BTS-DSN can produce a vessel probability map that has the same size as an input retinal image by a single forward propagation. Meanwhile, we use bottom-top short connections between side-output layers to pass low-level detail information to high-level layers. Also, the structural information of a high-level layer is passed back to a low-level layer by top-bottom short connection. We use VGGNet \cite{Simonyan2014Very} and ResNet-101 \cite{he2016deep} as backbone network respectively, and experimental results show that short connections can work on both cases. In addition, VGGNet behaves better than ResNet-101 due to higher resolution used.
Extensive experiments on DRIVE, STARE and CHASE\_DB1 datasets have demonstrated competitive performance of the BTS-DSN. Specially, with the patch-level input, the BTS-DSN achieved 0.9806 AUC, and 0.8249 F1-score on DRIVE dataset, and resulted in 0.9859 AUC and 0.8421 F1-score on STARE dataset. With the image-level input, the BTS-DSN achieved 0.9840 AUC and 0.7983 F1-score on CHASE\_DB1 dataset.

The contributions of this paper are as follows.
\begin{itemize}
\item We propose a deeply-supervised fully convolutional neural network with bottom-top and top-bottom short connections (BTS-DSN) for vessel segmentation. On one hand, it employs deep supervision to improve model performance. On the other hand, the BTS-DSN can refine results in high-level side-outputs by passing low-level detail information to a high-level layer, and reduce noises in low-level side-outputs by passing high-level structural information back to a low-level layer, thus leading to a better fusion result.
\item We used VGGNet and ResNet-101 as backbone and conducted extensive experiments on DRIVE, STARE and CHASE\_DB1 - three popular datasets for vessel segmentation, and the results show that short connections are compatible with both VGGNet and ResNet-101. Moreover, the proposed BTS-DSN with VGGNet as backbone achieved competitive performance over state-of-the-art .
\item We employed cross-training experiments to show the generalization of BTS-DSN. Experimental results show that BTS-DSN achieves better performance over other methods and this indicate it is suitable for real world application.
\end{itemize}

The remainder of this paper is organized as follows. The BTS-DSN is described in detail in Section 2. Section 3 gives experiments and analysis. A conclusion is drawn in Section 4.

\begin{figure*}[htbp]
\centering
\includegraphics[width=0.99\textwidth]{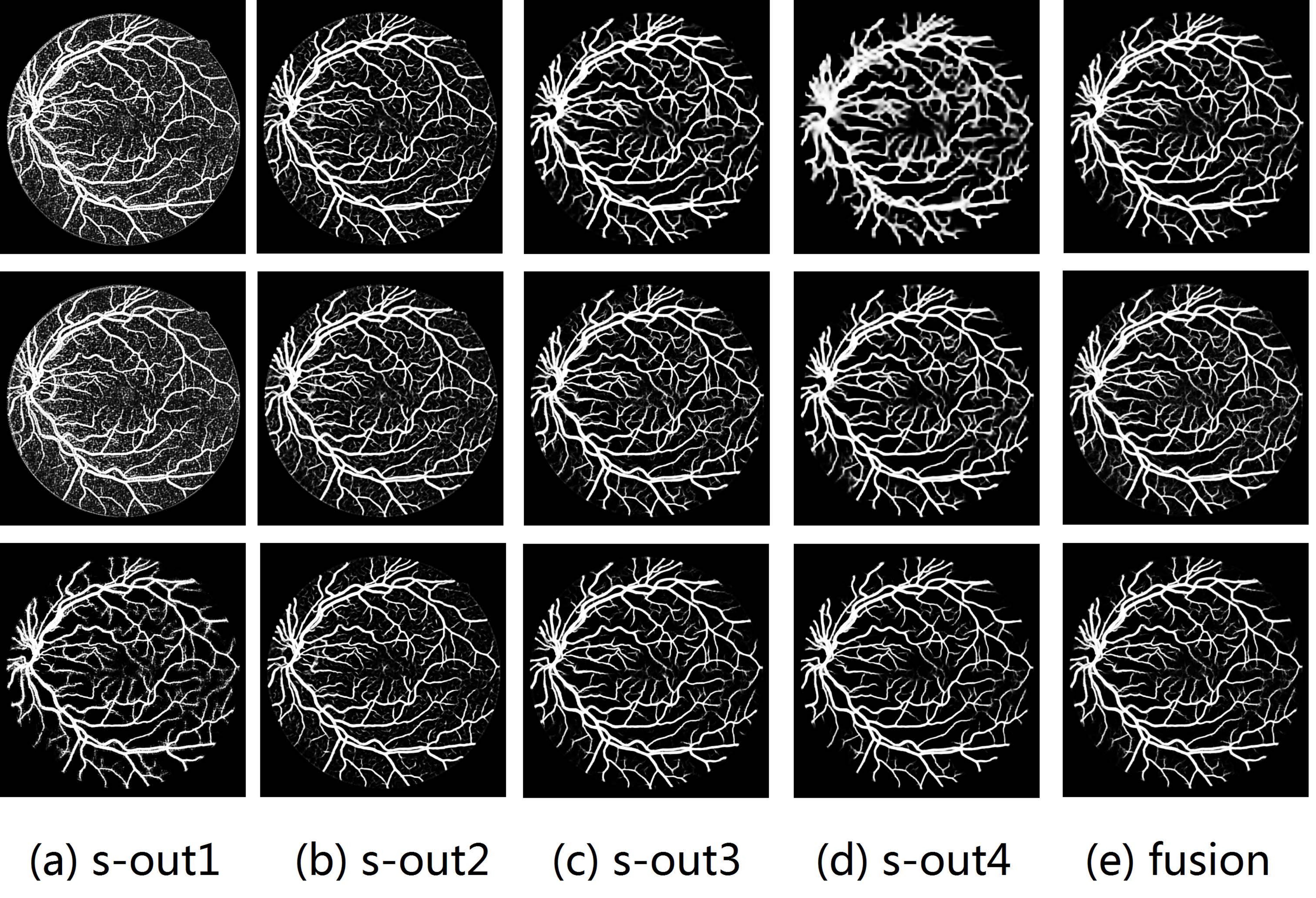}
\caption{Vessel segmentation results of side-output(s-out) layers produced by three networks. From top to bottom the network is normal DSN (with no short connections), BS-DSN (DSN with bottom-top short connections) and BTS-DSN (DSN with both bottom-top and top-bottom short connections), respectively.}
\label{fig:side-output}
\end{figure*}

\section{BTS-DSN}

As pointed out in recent works \cite{Xie2015Holistically,Long2017Fully}, a good semantic segmentation network should learn multi-level features. Further, it should have multiple stages with different receptive fields to learn more inherent features from different scales. FCN, taken as an example, uses skip connections to fuse multiple stages outputs, as well as the HED network, in which a series of side-output layers are added after each stage in VGGNet. The HED network was first proposed for edge detection, and further used for image-level vessel segmentation in recent studies \cite{Mo2017Multi,Fu2016DeepVessel}, with significant performance. However, our experimental results show that such network architecture is not appropriate for vessel segmentation directly. Figure~\ref{fig:side-output} provides such an illustration. Reasons for this phenomenon are straightforward. On one hand, the side-output of the first layer often contains too many noises. On the other hand, the features produced by the last side-output layer are too coarse due to information loss of pooling operation. Obviously, the inaccurate vessel map of side-output1 and side-output4 should have negative impacts on the final segmentation result.

\begin{figure}[!tb]
\centering
\includegraphics[width=0.9\textwidth]{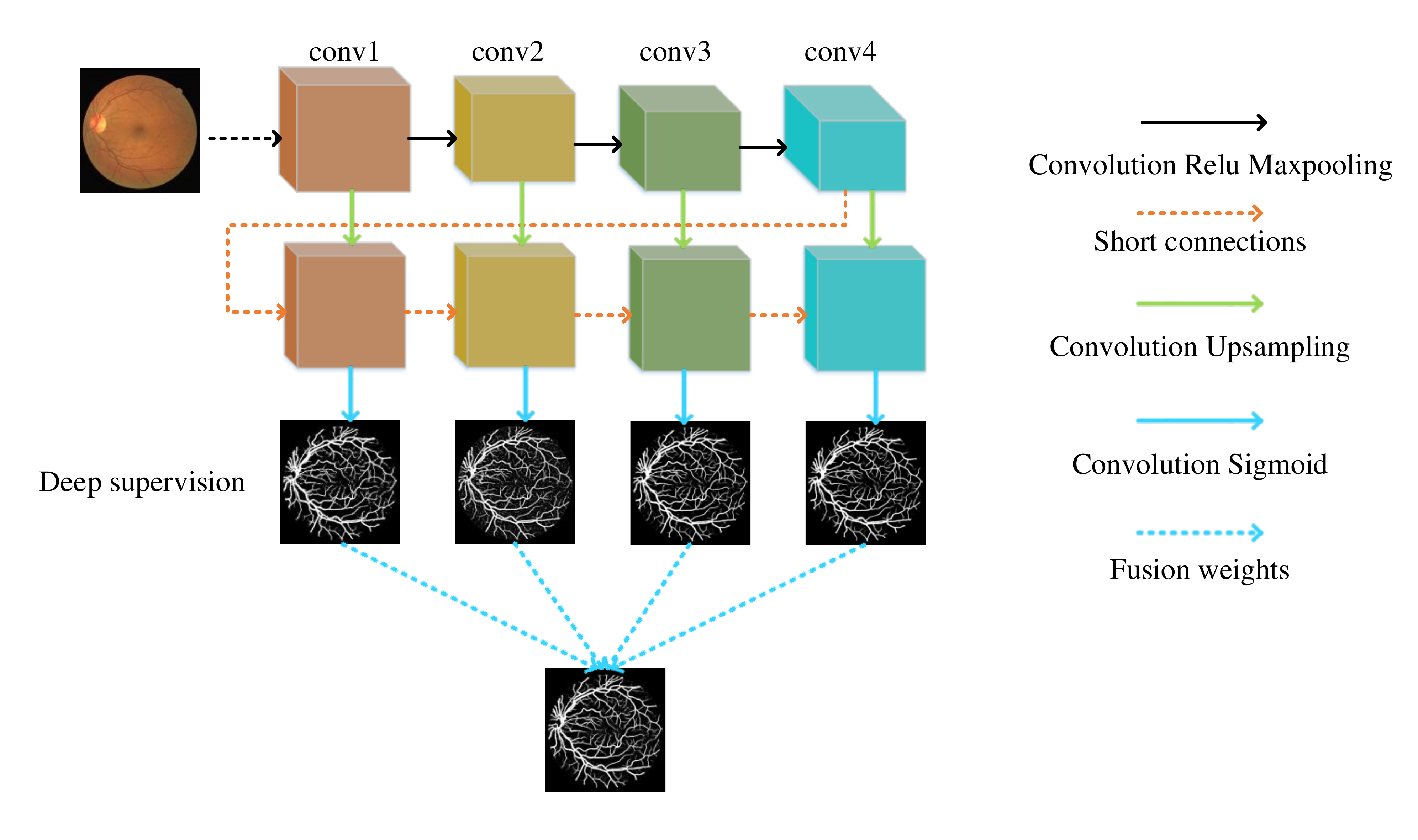}
\caption{An overview of the proposed BTS-DSN. Four uppermost cuboids represent four groups of convolutions of VGGNet. If ResNet-101 is used as backbone, then four uppermost cuboids are conv1, res2, res3 and res4.}
\label{fig:s-dsn-whole}
\end{figure}

To overcome the aforementioned problem, we propose a deeply supervised neural network with short connections for vessel segmentation. The overview of the BTS-DSN is illustrated in Figure~\ref{fig:s-dsn-whole}. Short connections are adopted to reduce noises of side-output1 as well as to make side-output4 less blurry. The following subsections will describe the BTS-DSN in details.

\subsection{Deep supervision}

\begin{figure}[tbhp]
\centering
\subfigure[HED]{
\includegraphics[width=0.14\textwidth]{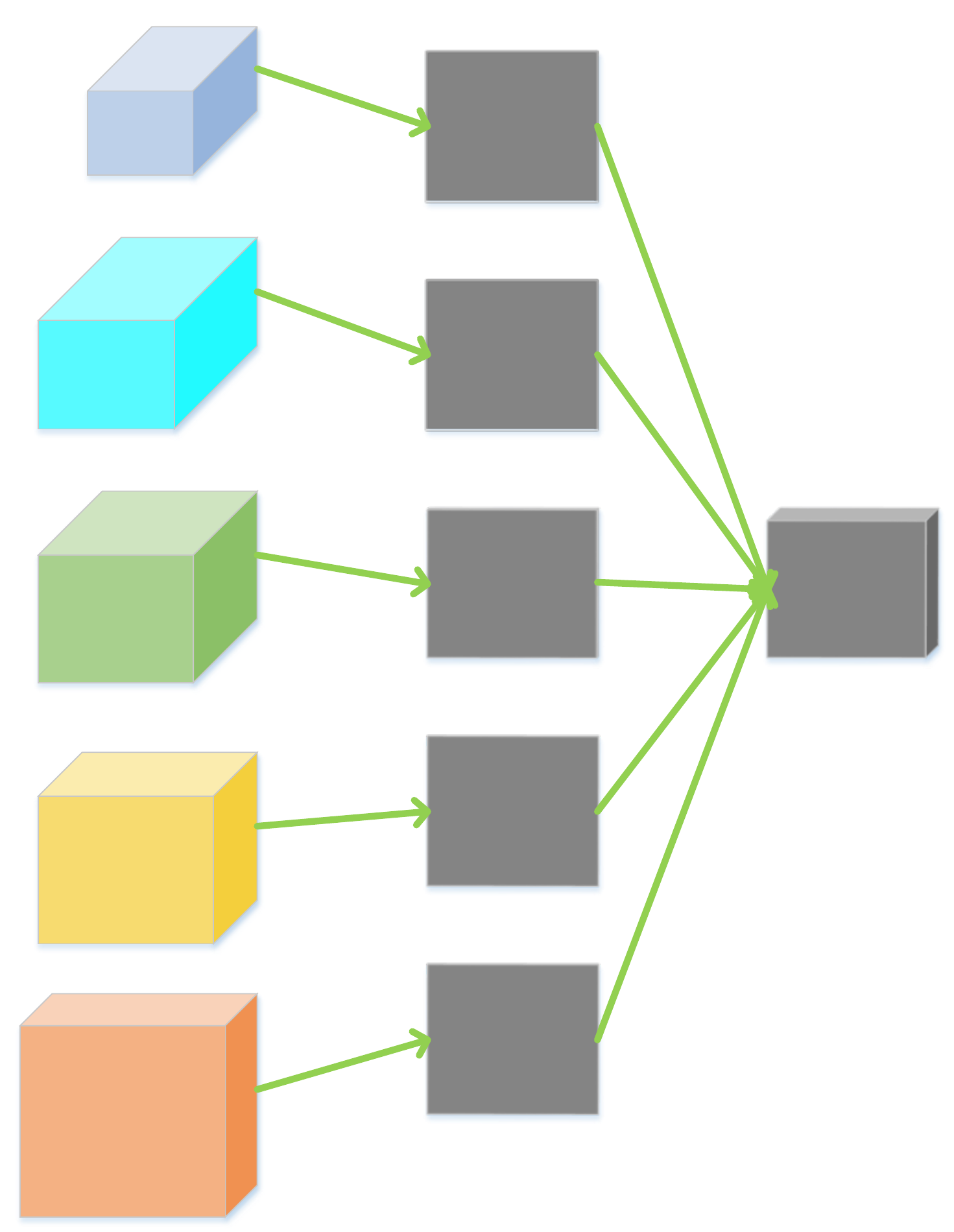}
\label{fig:hed}
}
\subfigure[DSN]{
\includegraphics[width=0.22\textwidth]{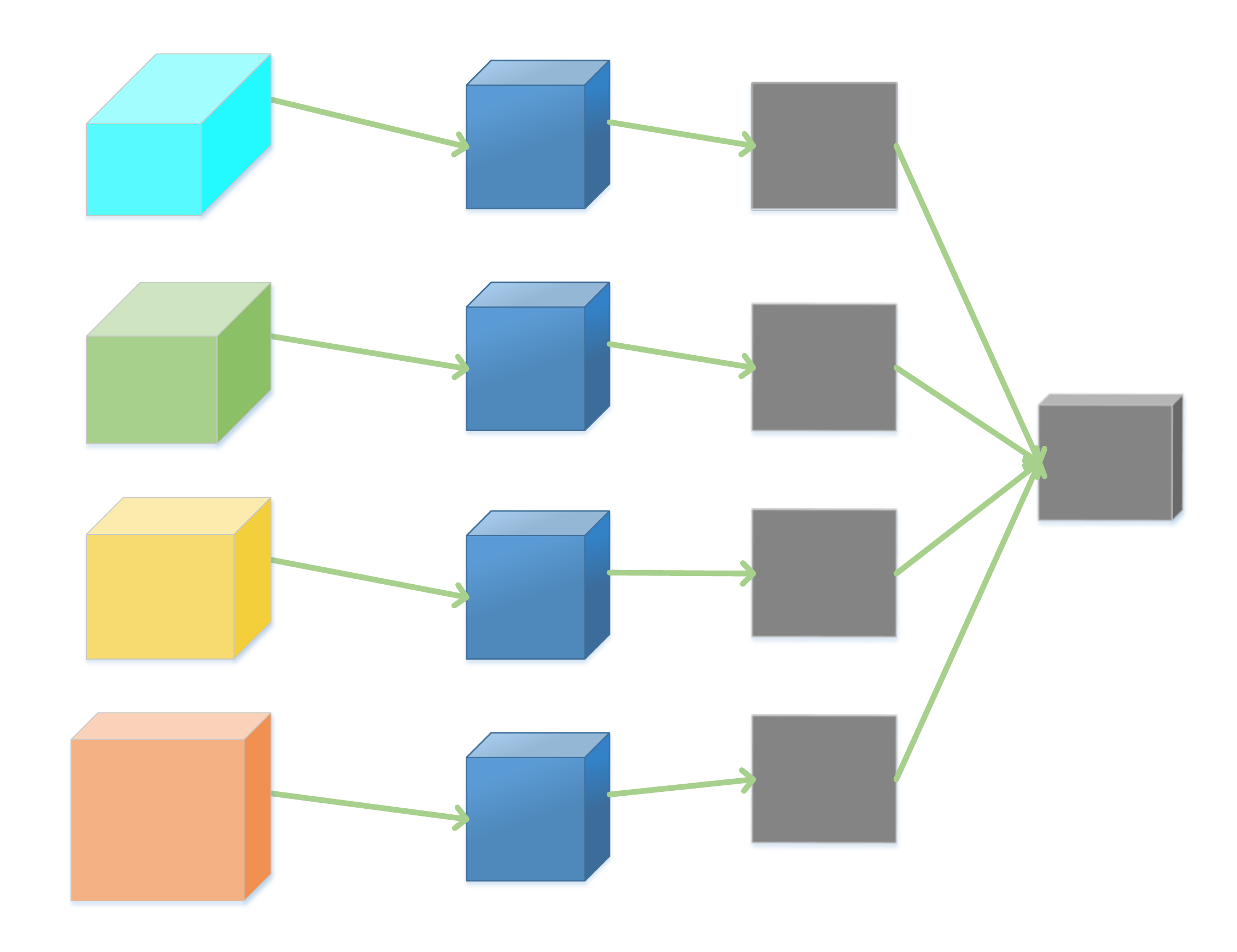}
\label{fig:dsn}
}
\subfigure[BS-DSN] {
\includegraphics[width=0.22\textwidth]{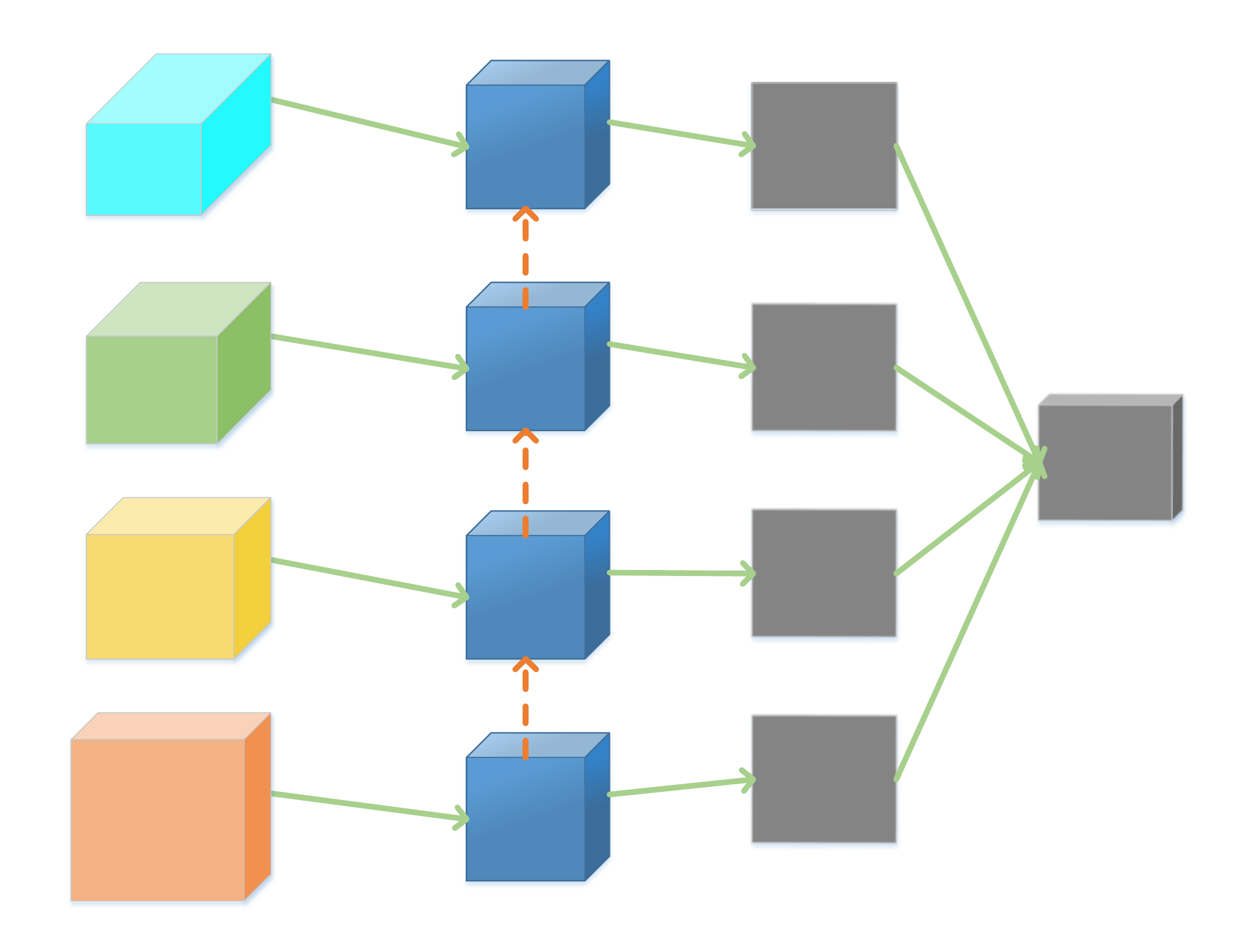}
\label{fig:fsdsn}
}
\subfigure[BTS-DSN] {
\includegraphics[width=0.3\textwidth]{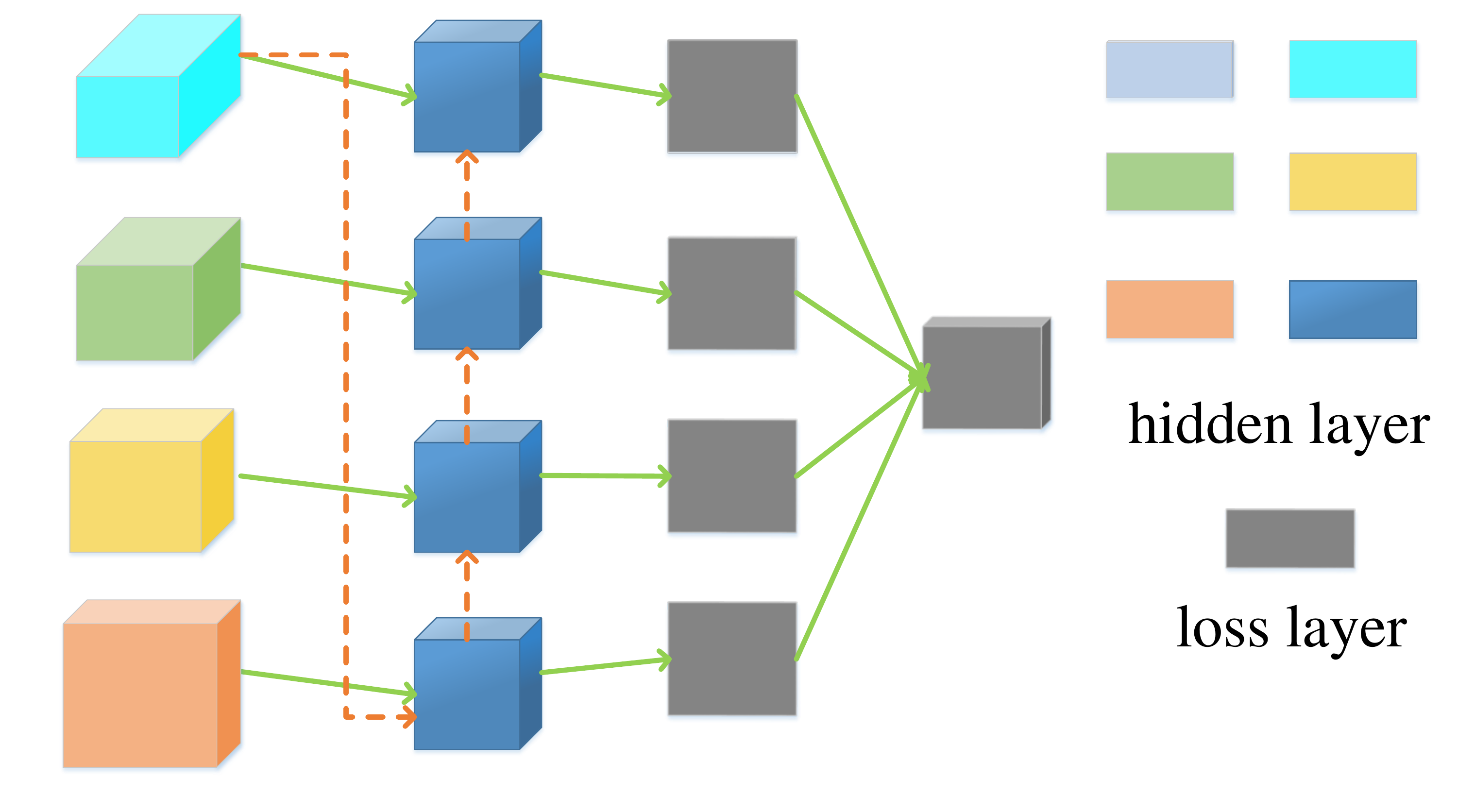}
\label{fig:sdsn}
}\caption{Network architectures of HED, DSN, BS-DSN and BTS-DSN. (a) HED with five side-output layers; (b) DSN with four side-output layers; (c) BS-DSN (DSN with bottom-top short connections); (d) BTS-DSN (BS-DSN with top-bottom short connection)}
\label{fig:hed-vs-sdsn}
\end{figure}

As is well known, it is hard to optimize a deep neural network due to the gradient vanish problem
\cite{Glorot2010Understanding}. To alleviate the gradient vanish problem and obtain a good vessel map, we use deep supervision information in the BTS-DSN.
Figure~\ref{fig:hed-vs-sdsn} gives an illustration of HED, DSN, BS-DSN and BTS-DSN. We can observe that DSN is based on HED except that there are only four side-output layers, and extra hidden layers are added in DSN for better utilizing deep supervision information. When bottom-top short connections are added to DSN, we get BS-DSN. Further, when top-bottom short connection is added to BS-DSN, we get BTS-DSN.

Let $S=\{(X_n, Y_n), n = 1, ..., N\}$, where $X_n=\{x_j^{(n)}, j = 1, ..., |X_n|\}$ denotes a raw input retinal image and $Y_n = \{y_j^{(n)}, j = 1, ..., |X_n|\}$, $y_j^{(n)}\in \{0,1\}$ denotes a corresponding ground truth binary vessel map for image $X_n$. Since each image is processed independently, subsequently the subscript $n$ is omitted for national convenience. For convenience, we denote the collection of all convolutional layer parameter of the standard network as $W$, and assume that there are $M$ side-output layers in the network. Each side-output layer can be regarded as a stand-alone image-level classifier, in which corresponding weights are denoted as $w=(w^{(1)},...,w^{(M)})$. The objective function of the side-output layer is formulated as follows:
\begin{align}
  L_{side}(W,w) = \sum_{m=1}^M \alpha_ml_{side}^{(m)}(W,w^{(m)})
\end{align}
where $\alpha_{m}$ is the weight of the $m$-th side-output layer and $l_{side}^{(m)}$ denotes the image-level loss function for side-output $m$. Since the distribution of vessel/non-vessel pixels is heavily biased and nearly 90\% of the pixels are non-vessel for a retinal image, thus a class-balanced cross-entropy loss function is used.
\begin{align}
l_{side}^{(m)}(W,w^{(m)}) &= -\frac{|Y_-|}{|Y|}\sum\limits_{j\in Y_+}\log Pr(y_j = 1|X;W,w^{(m)}) \nonumber \\
&- \frac{|Y_+|}{|Y|}\sum\limits_{j\in Y_-}\log Pr(y_j = 0|X;W,w^{(m)})
\end{align}

where $|Y_-|$ and $|Y_+|$ denote non-vessel and vessel pixels in a ground truth $Y$, respectively. $Pr(y_j=1|X;W,w^{(m)})$ is computed by a sigmoid function.

To utilize each side-output vessel probability map, a weight-fusion layer is adopted. The fusion output $Y_{fuse}$ is defined as follows:

\begin{align}
  Y_{fuse}= \sigma(\sum_{m=1}^{M}h_m{Y_{side}^{(m)}})
\label{equ:yfuse}
\end{align}

where $\sigma(\cdot)$ is a sigmoid function, $h_m$ is a fusion weight, and $Y_{side}^{(m)}$ is a vessel probability map of side-output $m$.
The side-output loss functions and the fusion-weight layer loss function are optimized together by back-propagation algorithm.

\begin{figure}
\centering
\includegraphics[width=0.95\textwidth]{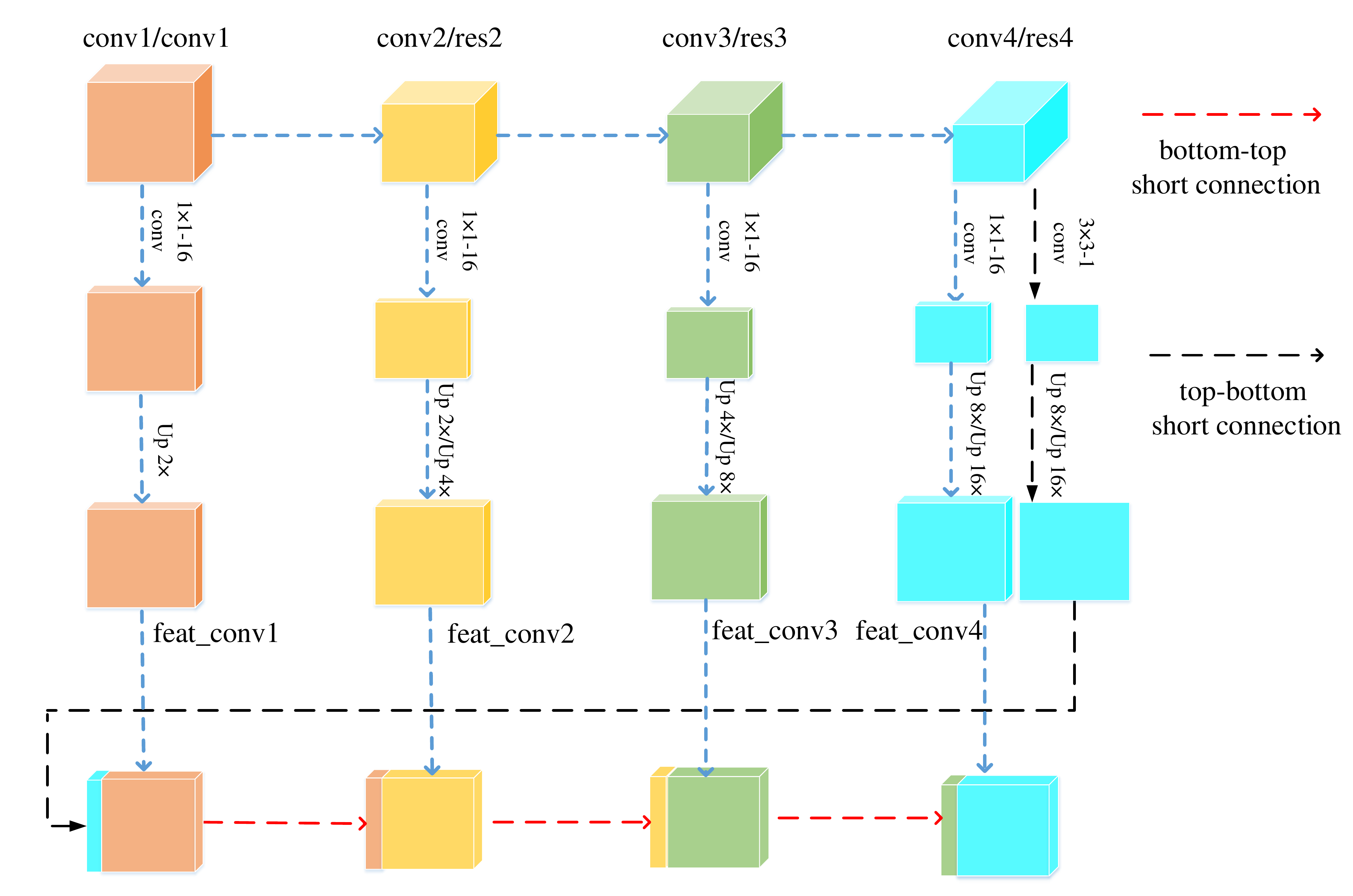}
\caption{An illustration of short connections. We use four groups of convolution for VGGNet and ResNet-101.}
\label{fig:short-connect}
\end{figure}

\begin{figure}[tbhp]
\centering
\subfigure[]{
\includegraphics[width=0.45\textwidth]{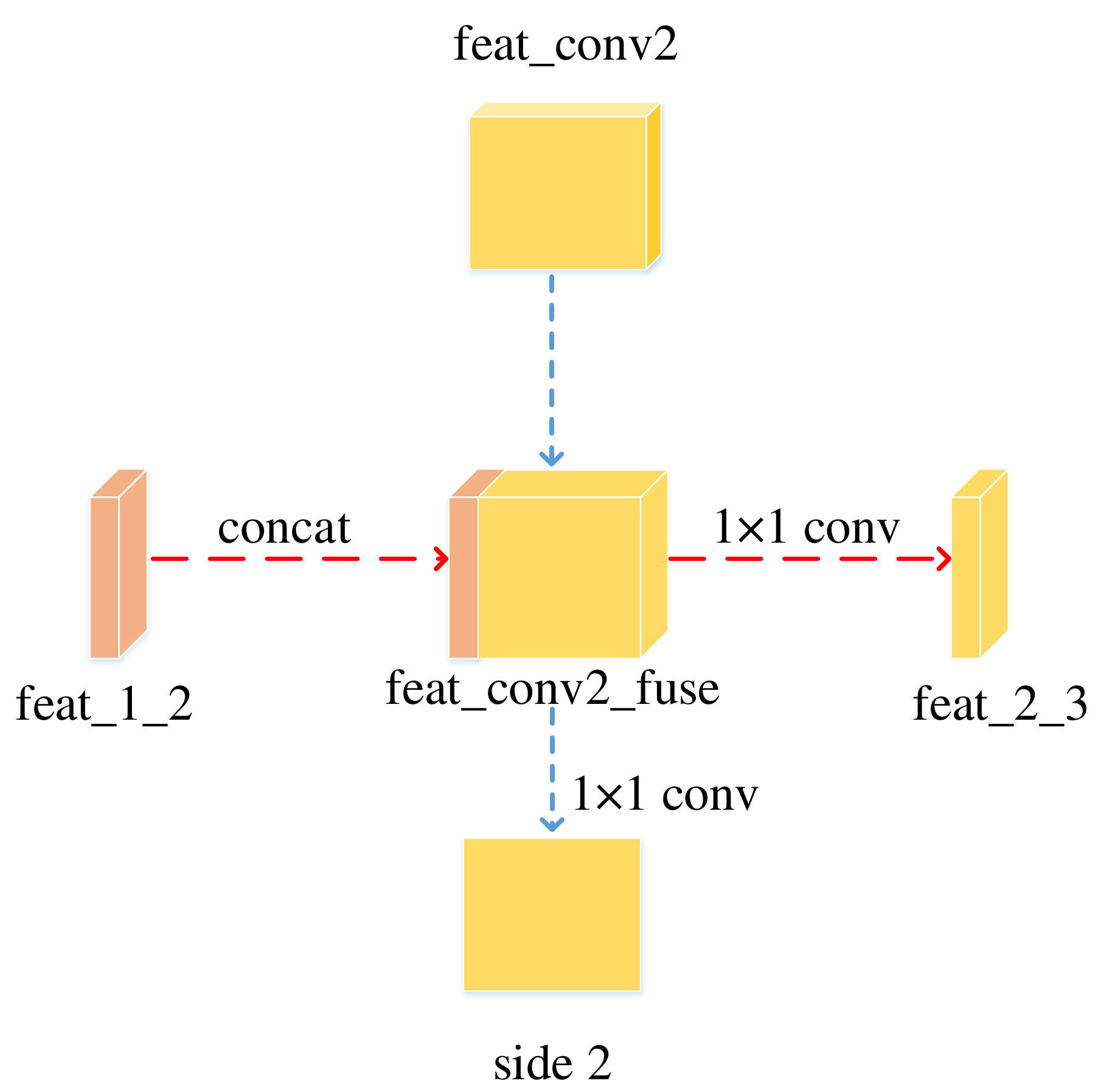}
\label{fig:for12}
}
\subfigure[]{
\includegraphics[width=0.415\textwidth]{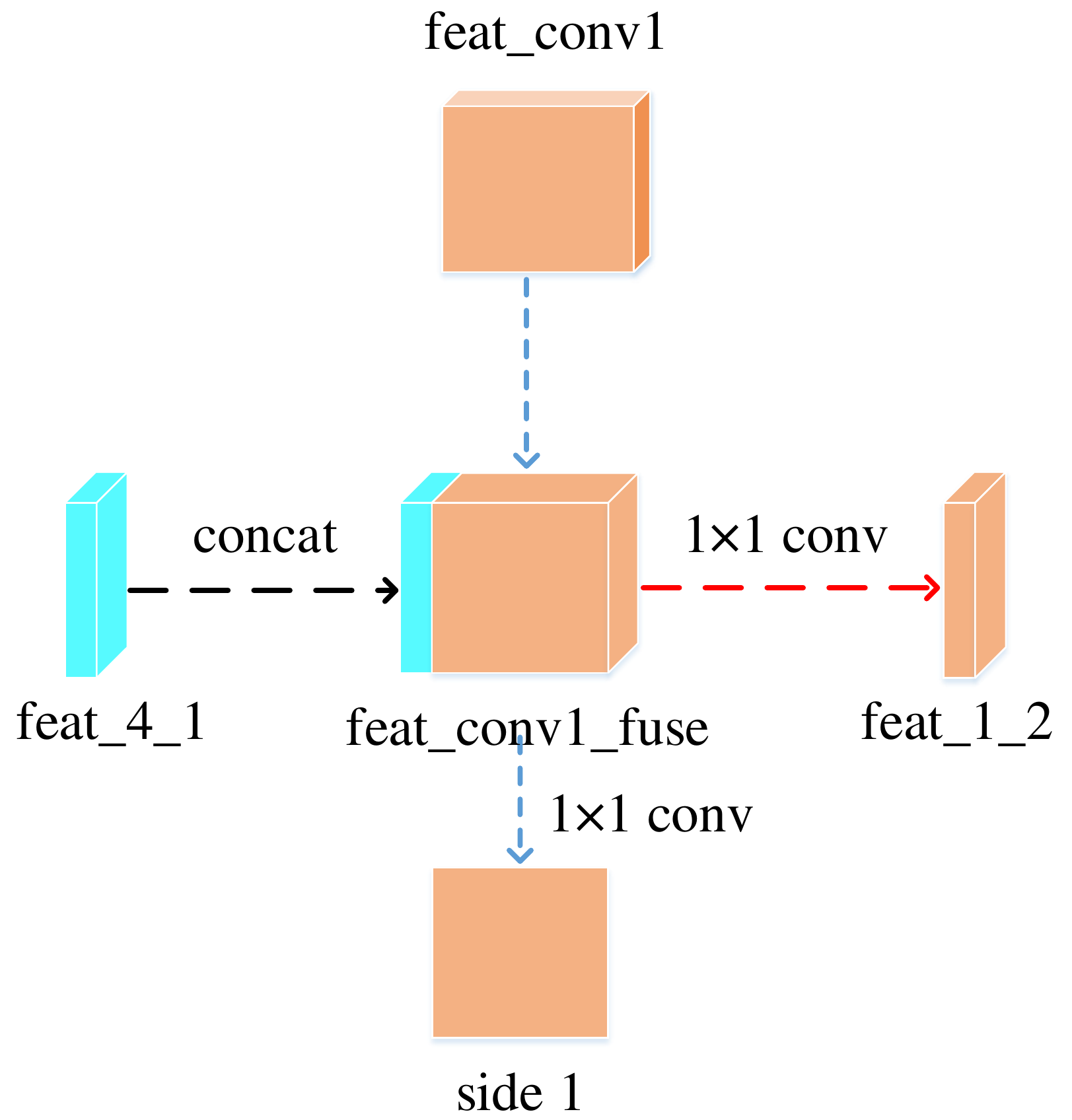}
\label{fig:back41}
}\caption{(a) Bottom-top short connection between feat\_conv1 and feat\_conv2 (b)top-bottom short connection between feat\_conv4 and feat\_conv1}
\label{fig:for12_back41}
\end{figure}

\subsection{Bottom-top short connections}
With the deepening of DSN network, the receptive field of each side-output layer gets larger, which makes the corresponding vessel map much blurrier as observed from the first row in Figure~\ref{fig:side-output}, especially for side-output4. These observations inspired us to pass low level fine semantic information to high levels to alleviate the blurring situation.

We adopted bottom-top short connections to deliver detail information just as shown in Figure~\ref{fig:short-connect}.
There are three bottom-top short connections in total. Suppose we use VGGNet as backbone. We can observe from Figure~\ref{fig:short-connect} that there are four groups of convolution (conv1, conv2, conv3 and conv4) for feature learning in total. We first convolved the last convolution of each group using 16 convolution kernels with size 1$\times$1. Then, the obtained feature maps are up-sampled 1$\times$, 2$\times$, 4$\times$, 8$\times$ respectively to restore to original resolution.
Bottom-top short connections are among feat\_conv1, feat\_conv2, feat\_conv3, and feat\_conv4.

Let's take bottom-top short connection between feat\_conv1 and feat\_conv2 as an example (see Figure~\ref{fig:for12}).
The information (feat\_1\_2) passed from feat\_conv1 is concatenated with feat\_conv2 to get feat\_conv2\_fuse. Then, one hand hand, we perform a 1$\times$1 convolution operation on feat\_conv2\_fuse to get the information (feat\_2\_3) delivered to feat\_conv3. On the other hand, we performed convolution operation with a kernel size of 1$\times$1 and sigmoid transformation for feat\_conv2\_fuse sequentially to obtain the segmentation result (side 2). At last, side 2 is compared with the ground truth to get the loss of the second side-output layer.

\subsection{Top-bottom short connection}
Bottom-top short connections aim to refine high-level segmentation results. However, we can observe from the first two rows in Figure~\ref{fig:side-output} that the vessel map generated by the first side-output layer contains too many noises while the map generated by the last side-output could capture the main vessel structure. Therefore, we propose delivering high-level structural information to the first side-output layer to reduce its noises. We implemented this kind of information delivery by a top-bottom short connection from conv4 to feat\_conv1, which can been seen in Figure~\ref{fig:short-connect}. We first convolved the last convolution of conv4 using 1 convolution kernels with size 3$\times$3. Then the obtained feature map are up-sampled 8$\times$ to get feat\_4\_1. The information (feat\_4\_1) passed from conv4 are concatenated with feat\_conv1 to form feat\_conv1\_fuse (see Figure~\ref{fig:back41}). At last, one hand hand, we perform a 1$\times$1
convolution operation on feat\_conv1\_fuse to get the information (feat\_1\_2) delivered to feat\_conv2. On the other hand, we performed convolution operation with a kernel size of 1$\times$1 and sigmoid transformation for feat\_conv1\_fuse sequentially to obtain the segmentation result (side 1). At last, side 1 is compared with the ground truth to get the loss of the first side-output layer.

\subsection{Inference}
When it comes to the inference phase, given a retinal image, both side-output layers and weighted-fusion layer produce a vessel probability map. The output of the weighted-fusion layer is often regarded as the final vessel segmentation result because it fuses multi-scale vessel maps together. The output of BTS-DSN is defined as follows:

\begin{align}
Y_{BTS\textrm{-}DSN} = Y_{fuse} = \sigma(\sum_{m=1}^{M}h_m{Y_{side}^{(m)}})
\label{equ:ysdsn}
\end{align}

We set M to 4 in our experiments. The meanings of the parameters in Equation~\ref{equ:ysdsn} are the same as those in Equation~\ref{equ:yfuse}.

\section{Experiments and analysis}
In this section, we will describe datasets used, evaluation criteria for vessel segmentation, implementation details of BTS-DSN, and experimental results and analysis.
\subsection{Materials}
We have evaluated our method on three publicly available retinal image vessel segmentation datasets: DRIVE, STARE and CHASE\_DB1.

The DRIVE \cite{staal2004ridge} dataset consists of 40 color fundus photographs with a resolution 565$\times$584. The dataset has been divided into a training set and a test set, each containing 20 images. For training images, a single manual segmentation of the vasculature is available. For test cases, two manual segmentations are available. One is used as the gold standard, and the other one can be used to compare computer generated segmentations with those by an independent human observer. For each image in DRIVE, a binary mask for FOV (Field of View) area is provided. To select the best epoch, we divide training set into training set and validation set. We use the first 15 images for training and the rest 5 images for validation.

The STARE \cite{stare} dataset consists of 20 fundus images with a resolution 700$\times$605. Each image has pixel-level vessel annotation provided by two experts. The annotations by the first expert are used as ground truth. In addition, this dataset does not provide partition of training set and testing set. Therefore, we used the same data partitioning method as in literatures \cite{driu,Fu2016DeepVessel,ricci2007retinal} (10 images for training and the rest 10 images for testing). Note that the same method of data partitioning is used in all the comparison methods in the paper. We select the first 7 images for training and the rest 3 images for validation.

The CHASE\_DB1 \cite {Fraz2012Blood} dataset contains 28 retinal images with a resolution 999$\times$960. The images are collected from both the left and right eyes of 14 people. Since the partition of training set and testing set is not present, we divided this dataset into 2 sets according to \cite{Li2016A,wu2018,jbhi2018,Azzopardi2015Trainable,Muhammad2012An}, where the first 20 images are considered as training set and the rest 8 images are considered as testing set.
Note that the same method of data partitioning is used in all the comparison methods in the paper.
In addition, due to the high resolution of retinal images in this dataset and limited GPU memory, the images are first scale to a much lower resolution so that it is possible to fed them into GPU memory. We divide the training set into two parts as well in this dataset. We select the first 15 images for training and the rest 5 images for validation.

\subsection{Implementation details}

\subsubsection{Data augmentation}
As there are fewer training images compared with the model complexity, training set augmentation methods are adopted. We used various transformations to augment the training set, including rotation, flipping and scaling. The training images were augmented by a factor of 13, 16 and 40 on DRIVE, CHASE\_DB1 and STARE, respectively.

\subsubsection{Model implementation}
We built the BTS-DSN architecture based on a publicly available convolutional network framework Caffe \cite{Jia2014Caffe}. Short connections of the BTS-DSN can be directly implemented by using the concatenate layer in Caffe.
\subsubsection{Parameter settings}
When the backbone is VGGNet, we fine-tuned our network with a learning rate of 1e-8, a weight decay of 0.0005, and a momentum of 0.9. We use a fixed learning rate. Since a fine retinal vessel is merely one pixel wide and too thin to respond in high layers, thus we took four side output layers in DSN, BS-DSN and BTS-DSN.
When the backbone is ResNet-101, we fine-tuned our network with a learning rate of 1e-7, a weight decay of 0.0005, and a momentum of 0.9. We used four groups convolution, res5 is dropped. We use validation set to select the best iterations on three datasets.
For bottom-top short connections, one feature map generated from a lower layer was concatenated to a higher layer. When training was conducted on the DRIVE and CHASE\_DB1 datasets, the input was color retinal images, and the output was corresponding vessel probability maps; no pre-processing or post-processing was performed. When training was conducted on the STARE dataset, we fed the green channel of fundus images into the BTS-DSN to enhance low-contrast vessels.

For patch-level S-DSN, we split a raw retinal image into 9 patches, each of which was 1/4 the size of the raw image. Further, patches were up-sampled 2$\times$. At last, the patches were used as the training samples to the patch-level BTS-DSN. When it came to the inference phase, we first get segmentation results of 9 patches and then they were puzzled together to form a whole map.

\subsubsection{Running environment}
The experimental platform we used in the study is a PC equipped with an Intel Xeon E5-2620 processor, an 128GB memory, and three NVIDIA GTX 1080ti GPUs, and the operating system is Ubuntu 16.04. It took about 0.2s to generate the final vessel map for image-level input on three datasets. As for patch-level input, it took about 2s to produce the vessel segmentation result.

\subsection{Evaluation criteria}
In vessel segmentation, each pixel belongs to a vessel or non-vessel pixel. By comparing the segmentation results with ground truth, we employed six evaluation criteria, including the area under the ROC curve (AUC), sensitivity (SE), specificity (SP), accuracy (ACC), F1-score and Matthews correlation coefficient (MCC). The evaluation measurements were calculated only for pixels inside the FOV area. The SE, SP, ACC, F1-score and MCC are defined as follows:

\begin{align}
SE=\frac{TP}{TP+FN}
\end{align}
\begin{align}
SP=\frac{TN}{TN+FP}
\end{align}
\begin{align}
ACC=\frac{TP+TN}{TP+FN+TN+FP}
\end{align}
\begin{align}
F1\textrm{-}score=\frac{2\times PR\times SE}{PR+SE}
\end{align}
\begin{align}
MCC=\frac{TP\times TN - FP\times FN}{\sqrt{(TP+FP)(TP+FN)(TN+FP)(TN+FN)}}
\end{align}

where $PR=\frac{TP}{TP+FP}$, the true positives (TP) are vessel pixels classified correctly, false negatives (FN) are vessel pixels mis-classified as non-vessel pixels, true negatives (TN) are non-vessel pixels classified correctly, and false positives (FP) are non-vessel pixels mis-classified as vessel pixels. The calculation of the SE, SP, ACC measurements is dependent on the threshold of the vessel probability map, but the AUC value takes into account of the fixed threshold. The ROC was plotted with the SE versus (1-SP) by varying the threshold. The perfect classifier achieved an AUC value equal to 1.

\subsection{Results}

\subsubsection{Ablation study}
To demonstrate effectiveness of the proposed short connections, we performed experiments on three different DSN architectures, namely DSN, BS-DSN and BTS-DSN. Moreover, we used VGGNet and ResNet-101 as backbone network respectively.
Tables~\ref{table:sdsn result_drive}, \ref{table:sdsn result_stare} and \ref{table:sdsn result_chasedb1} illustrates the comparison of statistical measures when the backbone is VGGNet.
Tables~\ref{table:sdsn result_drive_res}, \ref{table:sdsn result_stare_res} and \ref{table:sdsn result_chasedb1_res} illustrates the comparison of statistical measures when the backbone is ResNet-101.

We can observe from Tables~\ref{table:sdsn result_drive}, \ref{table:sdsn result_drive_res}, \ref{table:sdsn result_stare}, \ref{table:sdsn result_stare_res}, \ref{table:sdsn result_chasedb1} and \ref{table:sdsn result_chasedb1_res} that BS-DSN behaved better than DSN in SE, ACC, AUC, MCC and F1-score on all three datasets, which shows the effectiveness of bottom-top short connections. Moreover, compared with BS-DSN, BTS-DSN achieved much higher scores in terms of AUC, MCC and F1-score on all three datasets when the backbone is VGGNet, which demonstrate the effectiveness of top-bottom short connection. Specially, BTS-DSN achieved much higher scores in terms of five measures on DRIVE and STARE datasets, compared with BS-DSN.

In addition, we can observe from Figure~\ref{fig:side-output} that the side-output1 and side-output4 of the BTS-DSN were more accurate compared with those of the DSN.

\setlength{\tabcolsep}{2.6pt}
\begin{table*}
\begin{center}
\caption{Performance comparison of vessel segmentation results between HED and three different image-level DSN architectures with backbone VGGNet on DRIVE dataset (best results shown in bold)}
\begin{tabular}{ccccccc}
\hline
Model & SE & SP & ACC & AUC & MCC & F1-score \\
\hline
HED (Figure~\ref{fig:hed}) &0.7627 &0.9801 &0.9524 &0.9758 &0.7784 &0.8089 \\
DSN (Figure~\ref{fig:dsn}) &0.7716 &0.9797 &0.9532 &0.9778 &0.7830 &0.8125 \\
BS-DSN (Figure~\ref{fig:fsdsn}) &\bf0.7875 &0.9790 &0.9546 &\bf0.9796 &0.7904 &0.8189\\
BTS-DSN (Figure~\ref{fig:sdsn}) &0.7800 &\bf0.9806 &\bf0.9551 &\bf0.9796 &\bf0.7923 &\bf0.8208 \\
\hline
\end{tabular}
\label{table:sdsn result_drive}
\end{center}
\end{table*}

\setlength{\tabcolsep}{2.6pt}
\begin{table*}
\begin{center}
\caption{Performance comparison of vessel segmentation results among three different image-level DSN architectures with backbone ResNet-101 on DRIVE dataset (best results shown in bold)}
\begin{tabular}{ccccccc}
\hline
Model & SE & SP & ACC & AUC & MCC & F1-score \\
\hline
DSN (Figure~\ref{fig:dsn}) &0.7144 &0.9783 &0.9447 &0.9547 &0.7373 &0.7735 \\
BS-DSN (Figure~\ref{fig:fsdsn}) &0.7307 &\bf0.9790 &0.9474 &0.9580 &0.7528 &0.7853\\
BTS-DSN (Figure~\ref{fig:sdsn}) &\bf0.7535 &0.9787 &\bf0.9500 &\bf0.9678 &\bf0.7663 &\bf0.7987 \\
\hline
\end{tabular}
\label{table:sdsn result_drive_res}
\end{center}
\end{table*}

\setlength{\tabcolsep}{2.6pt}
\begin{table*}
\begin{center}
\caption{Performance comparison of vessel segmentation results between HED and three different image-level DSN architectures with backbone VGGNet on STARE dataset (best results shown in bold)}
\begin{tabular}{ccccccc}
\hline
Model & SE & SP & ACC & AUC & MCC & F1-score \\
\hline
HED (Figure~\ref{fig:hed}) &0.8076	&0.9822 &0.9641 &0.9824 &0.8019 &0.8268 \\
DSN (Figure~\ref{fig:dsn}) &0.8114 &\bf0.9829 &0.9651 &0.9864 &0.8105 &0.8320 \\
BS-DSN (Figure~\ref{fig:fsdsn}) &0.8184 &0.9824  &0.9655 &0.9866 &0.8110 &0.8340\\
BTS-DSN (Figure~\ref{fig:sdsn}) &\bf0.8201 &0.9828  &\bf0.9660 &\bf0.9872 &\bf0.8142 &\bf0.8362 \\
\hline
\end{tabular}
\label{table:sdsn result_stare}
\end{center}
\end{table*}

\setlength{\tabcolsep}{2.6pt}
\begin{table*}
\begin{center}
\caption{Performance comparison of vessel segmentation results among three different image-level DSN architectures with backbone ResNet-101 on STARE dataset (best results shown in bold)}
\begin{tabular}{ccccccc}
\hline
Model & SE & SP & ACC & AUC & MCC & F1-score \\
\hline
DSN (Figure~\ref{fig:dsn}) &0.7324 &\bf0.9850 &0.9589 &0.9668 &0.7666 &0.7933 \\
BS-DSN (Figure~\ref{fig:fsdsn}) &0.7554 &\bf0.9850  &0.9612 &0.9729 &0.7804 &0.8059\\
BTS-DSN (Figure~\ref{fig:sdsn}) &\bf0.7730 &0.9849 &\bf0.9630 &\bf0.9795 &\bf0.7923 &\bf0.8178 \\
\hline
\end{tabular}
\label{table:sdsn result_stare_res}
\end{center}
\end{table*}

\setlength{\tabcolsep}{2.6pt}
\begin{table*}
\begin{center}
\caption{Performance comparison of vessel segmentation results between HED and three different image-level DSN architectures with backbone VGGNet on CHASE\_DB1 dataset (best results shown in bold)}
\begin{tabular}{ccccccc}
\hline
Model & SE & SP & ACC & AUC & MCC & F1-score \\
\hline
HED (Figure~\ref{fig:hed}) &0.7516 &0.9805 &0.9597 &0.9796 &0.7489 &0.7815\\
DSN (Figure~\ref{fig:dsn}) &0.7735 &0.9806 &0.9618 &0.9831 &0.7653 &0.7924 \\
BS-DSN (Figure~\ref{fig:fsdsn}) &0.7809 &\bf0.9812  &\bf0.9630 &0.9838 &0.7729 &\bf0.7983\\
BTS-DSN (Figure~\ref{fig:sdsn}) &\bf0.7888 &0.9801  &0.9627 &\bf0.9840 &\bf0.7733 &\bf0.7983 \\
\hline
\end{tabular}
\label{table:sdsn result_chasedb1}
\end{center}
\end{table*}

\setlength{\tabcolsep}{2.6pt}
\begin{table*}
\begin{center}
\caption{Performance comparison of vessel segmentation results among three different image-level DSN architectures with backbone ResNet-101 on CHASE\_DB1 dataset (best results shown in bold)}
\begin{tabular}{ccccccc}
\hline
Model & SE & SP & ACC & AUC & MCC & F1-score \\
\hline
DSN (Figure~\ref{fig:dsn})      &0.7242 &\bf0.9802 &0.9570 &0.9735 &0.7308 &0.7641 \\
BS-DSN (Figure~\ref{fig:fsdsn}) &0.7362 &0.9796  &0.9575 &\bf0.9763 &0.7359 &0.7675\\
BTS-DSN (Figure~\ref{fig:sdsn})   &\bf0.7432 &0.9801  &\bf0.9586 &0.9738 &\bf0.7430 &\bf0.7705 \\
\hline
\end{tabular}
\label{table:sdsn result_chasedb1_res}
\end{center}
\end{table*}

\begin{figure}[hptb]
\centering
\subfigure{
\includegraphics[width=0.20\textwidth]{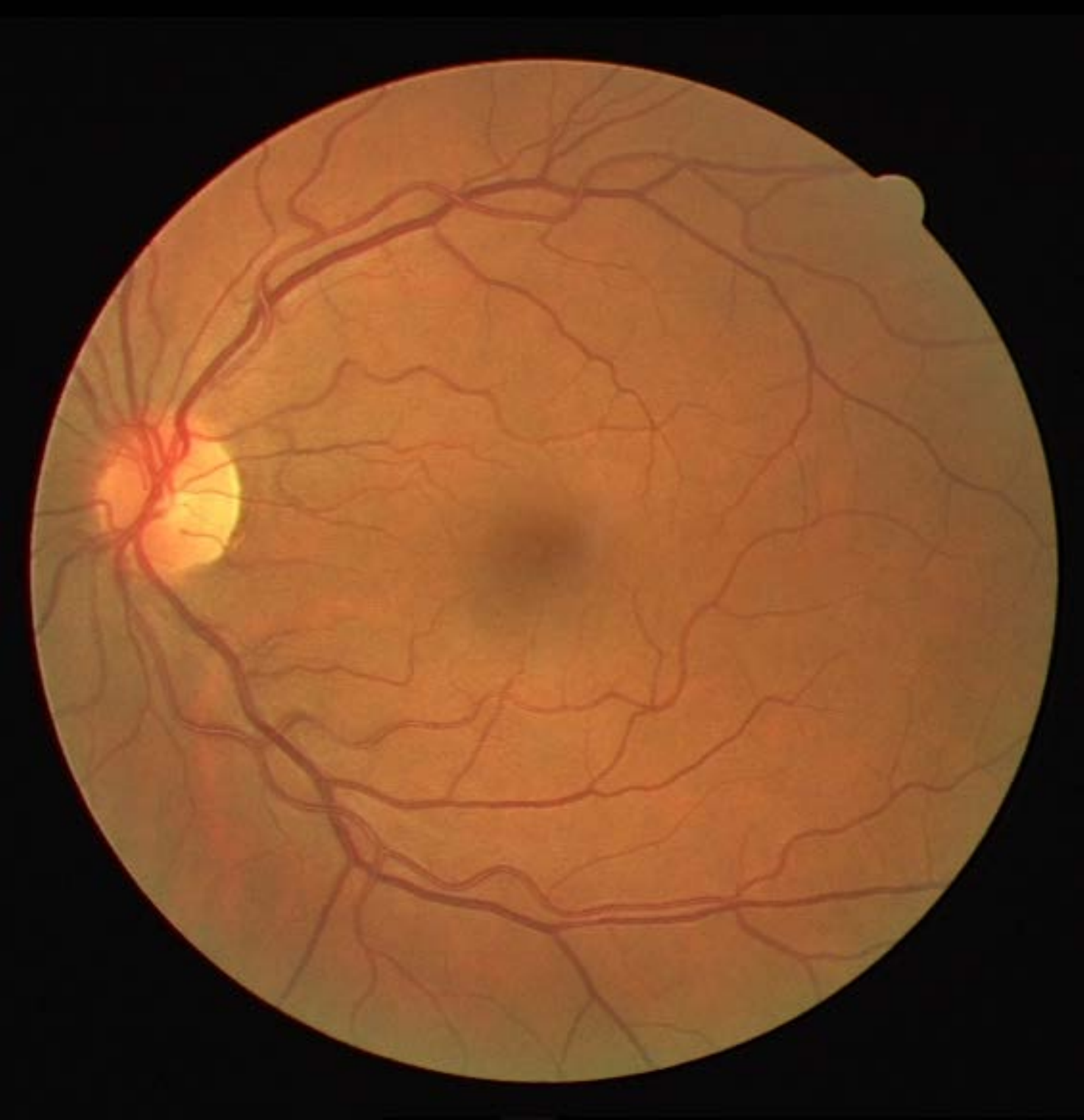}
}\subfigure {
\includegraphics[width=0.20\textwidth]{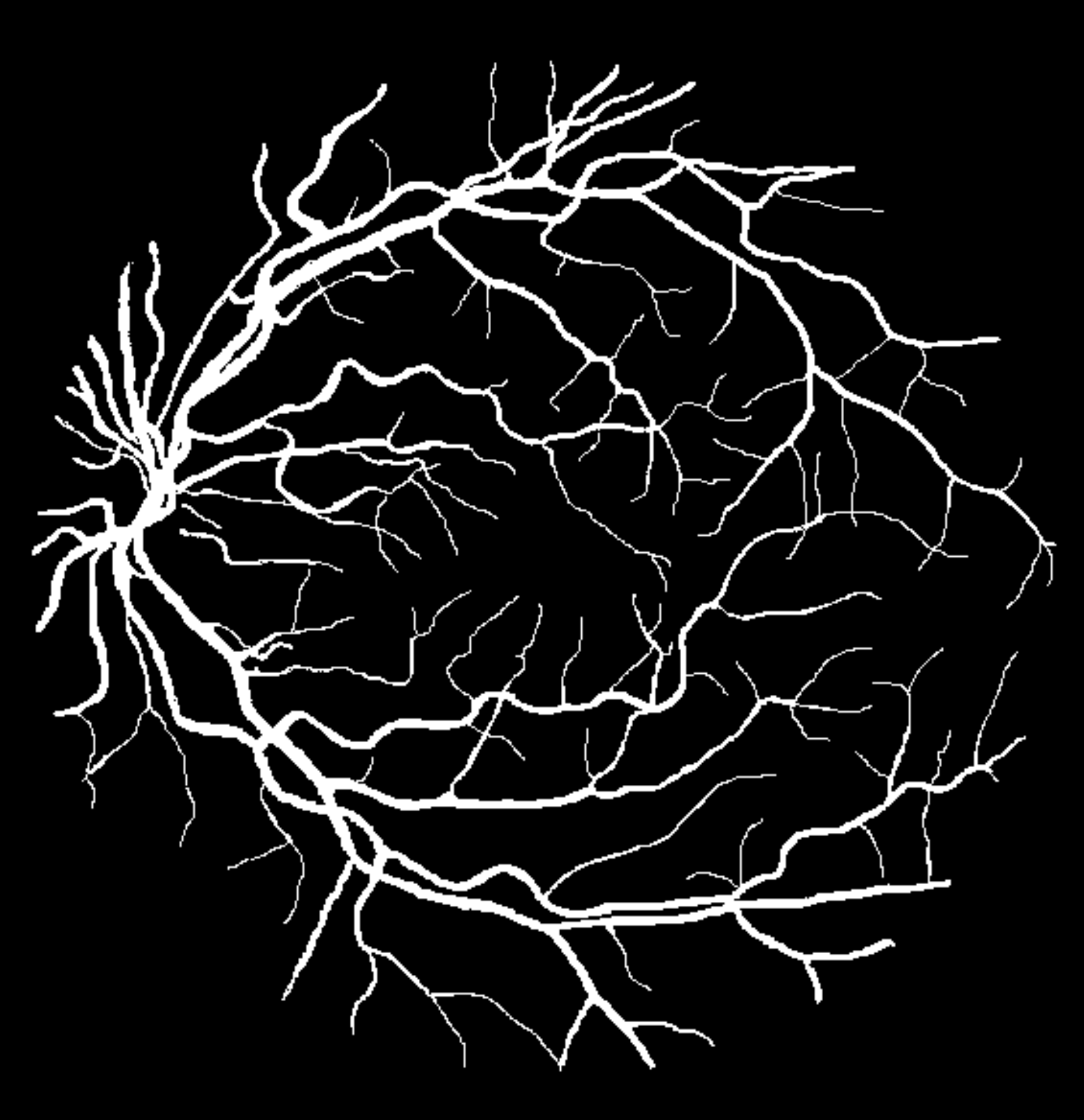}
}\subfigure{
\includegraphics[width=0.20\textwidth]{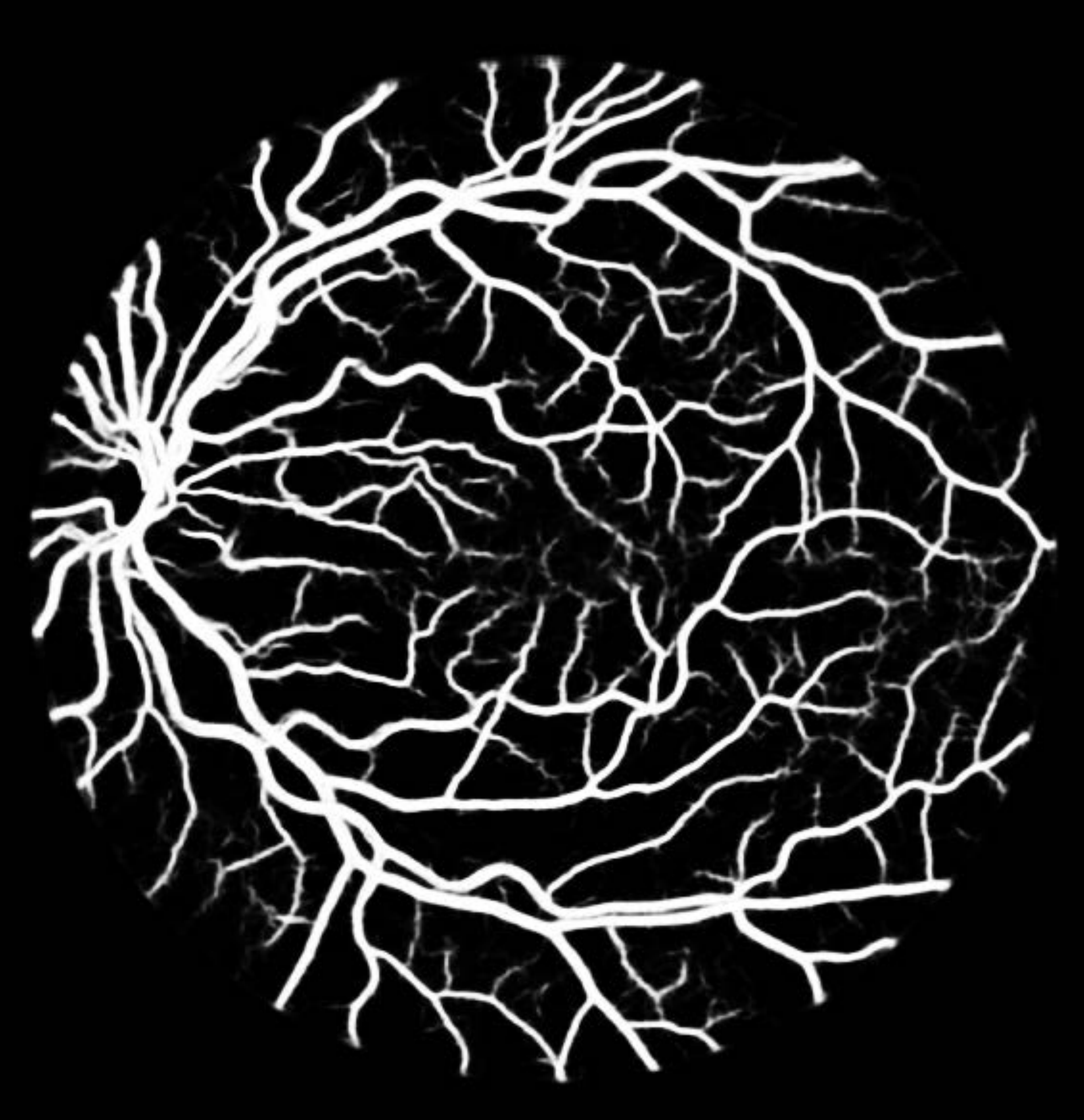}
}\subfigure{
\includegraphics[width=0.20\textwidth]{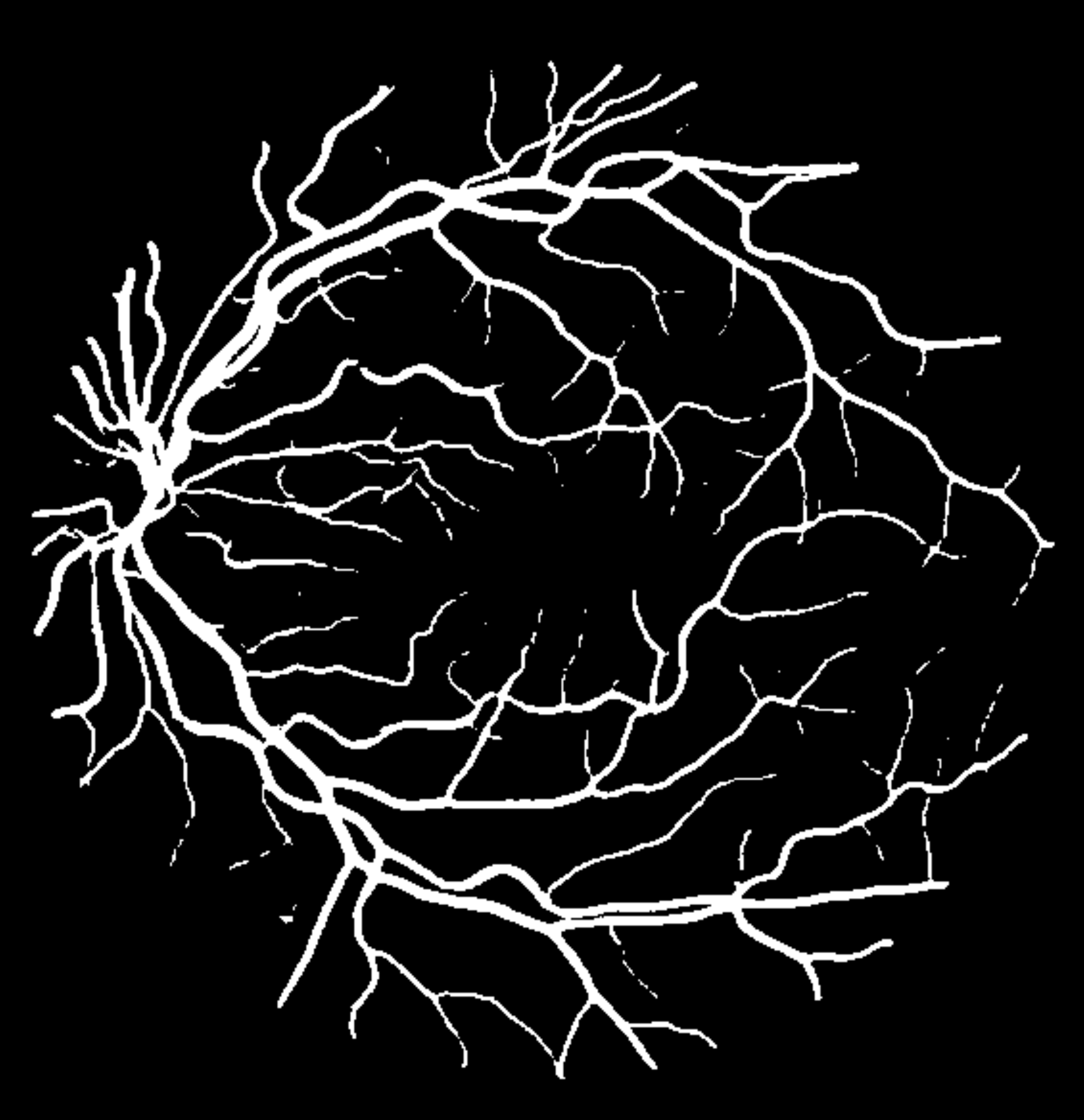}
}\\
\subfigure{
\includegraphics[width=0.20\textwidth]{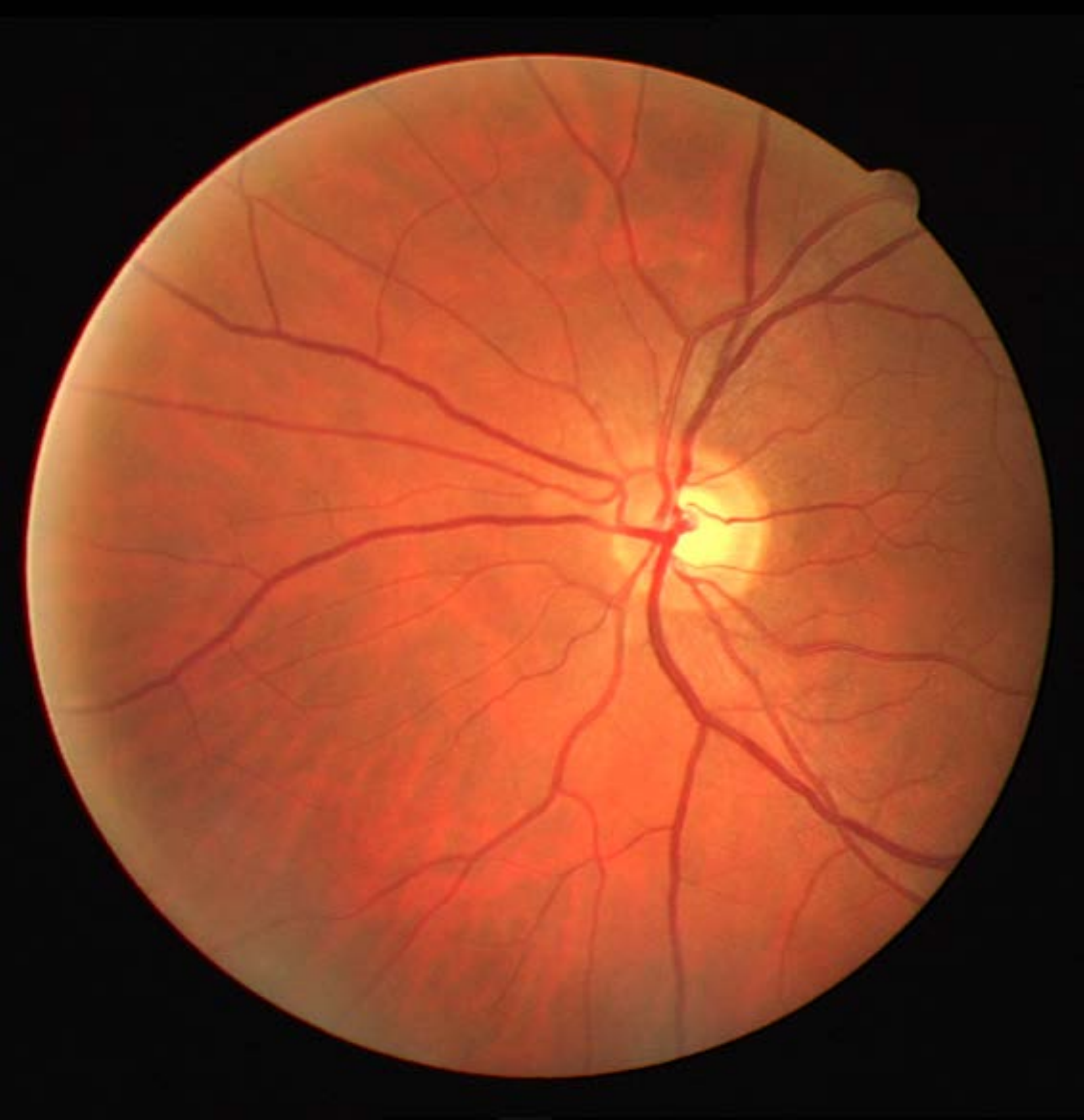}
}\subfigure {
\includegraphics[width=0.20\textwidth]{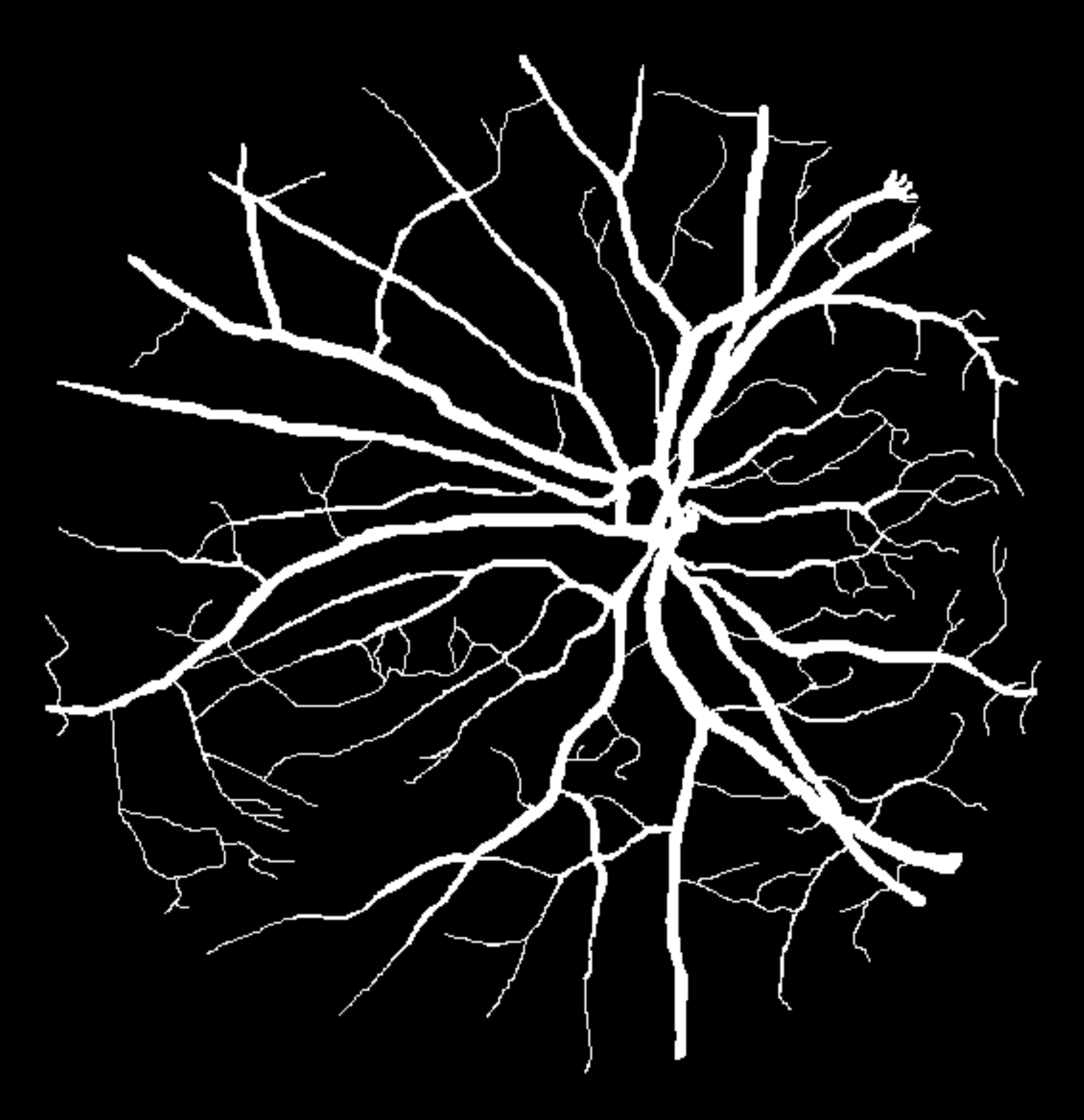}
}\subfigure{
\includegraphics[width=0.20\textwidth]{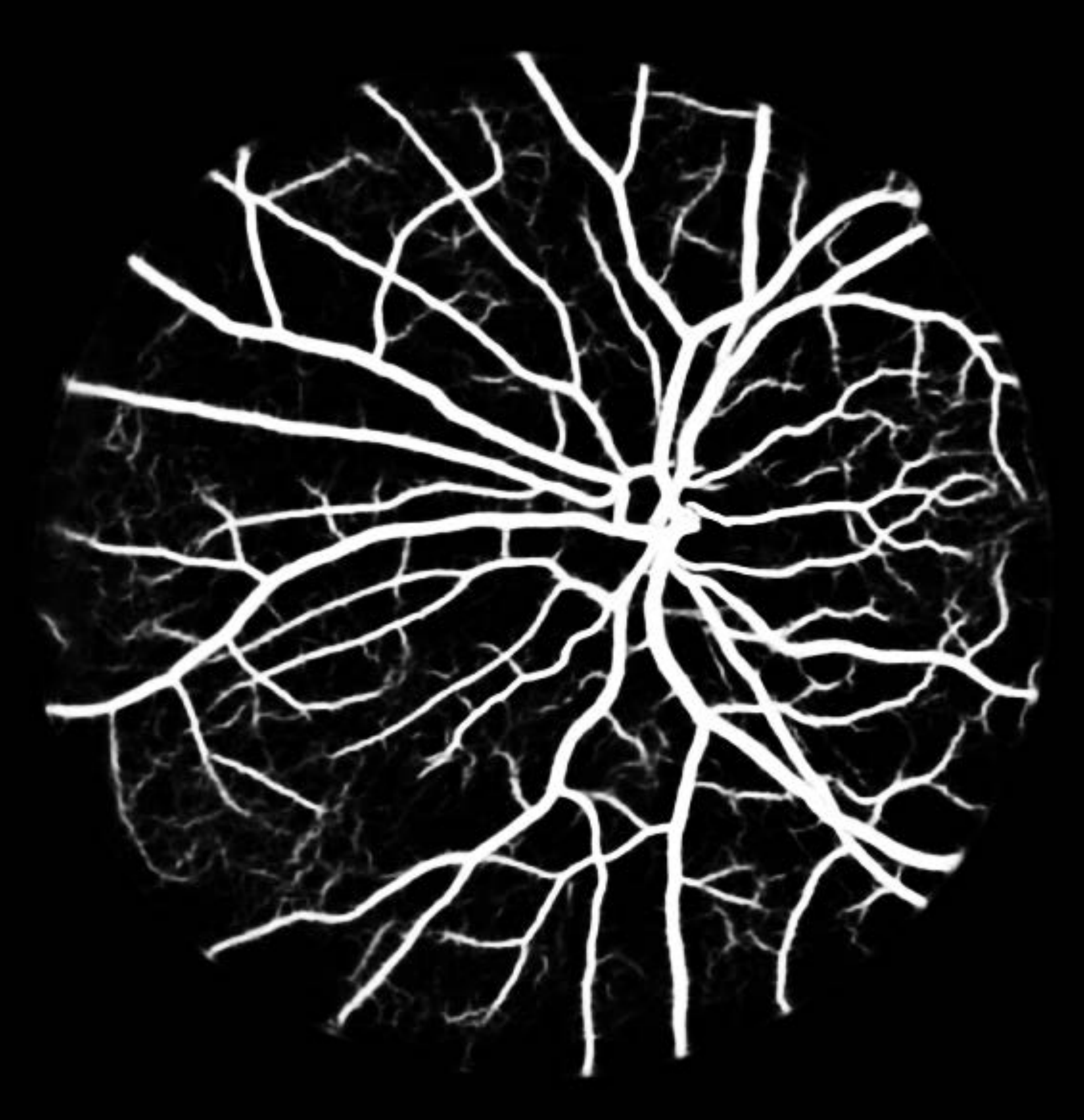}
}\subfigure{
\includegraphics[width=0.20\textwidth]{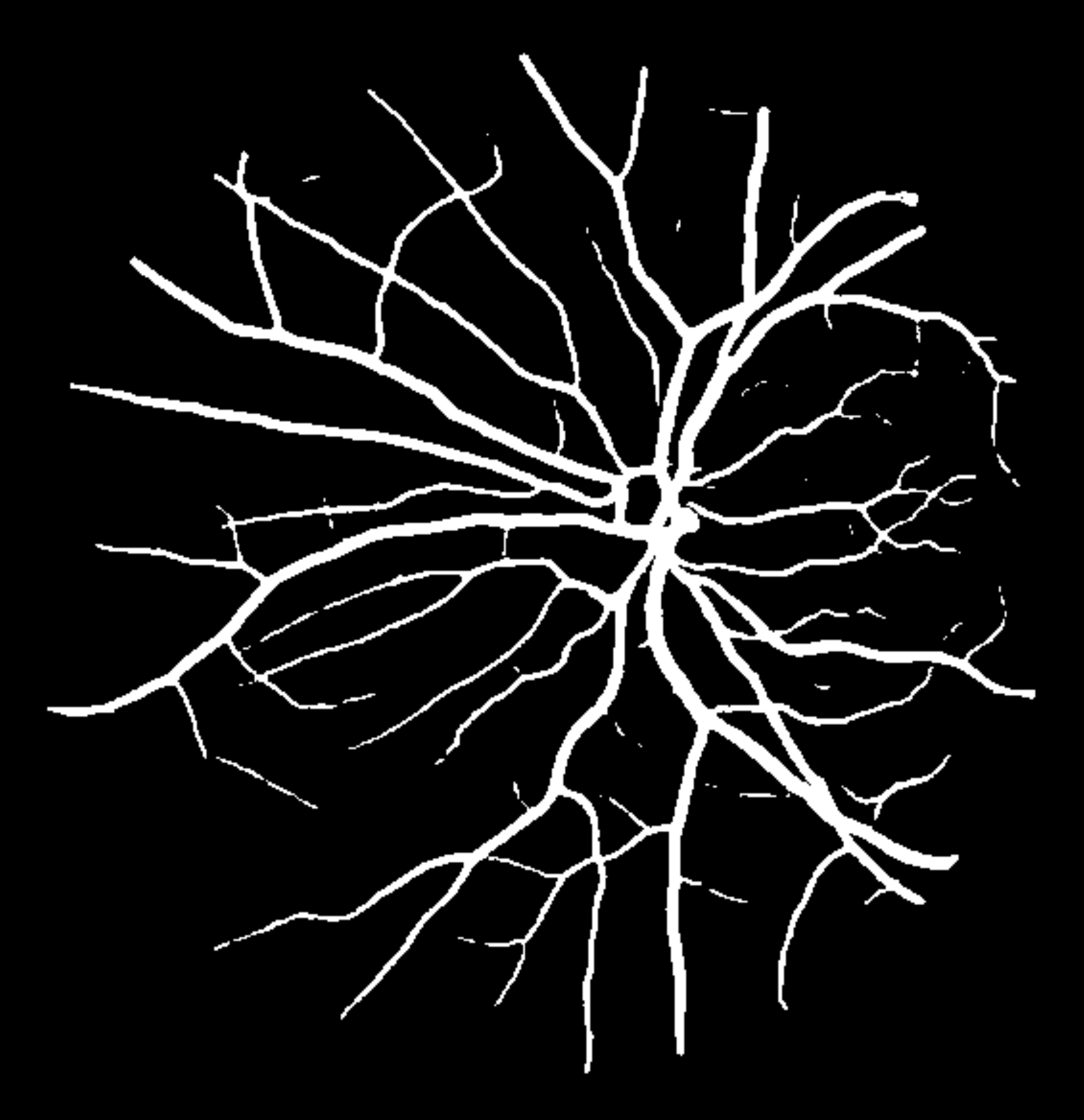}
}\\
\subfigure{
\includegraphics[width=0.20\textwidth]{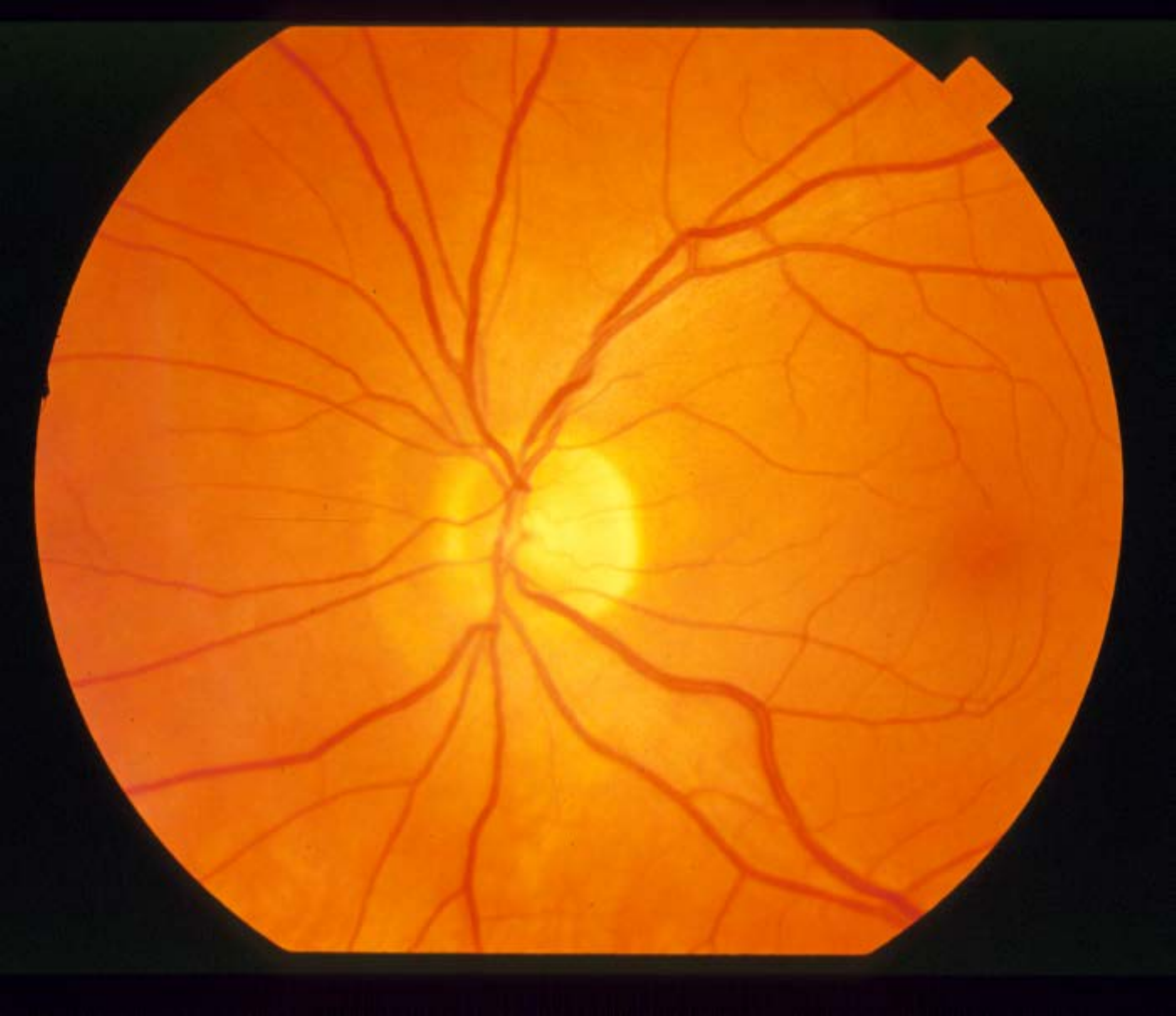}
}\subfigure {
\includegraphics[width=0.20\textwidth]{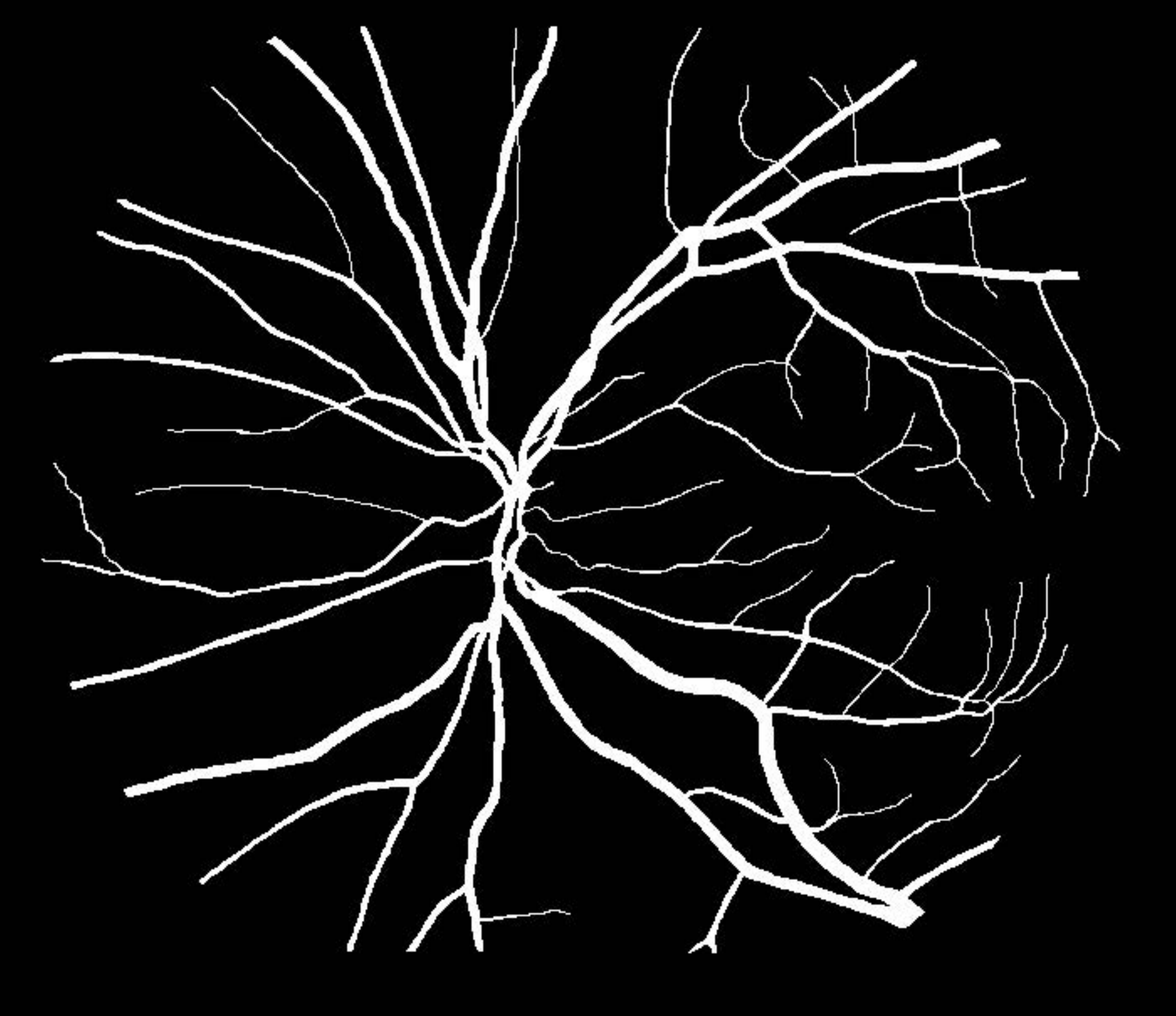}
}\subfigure{
\includegraphics[width=0.20\textwidth]{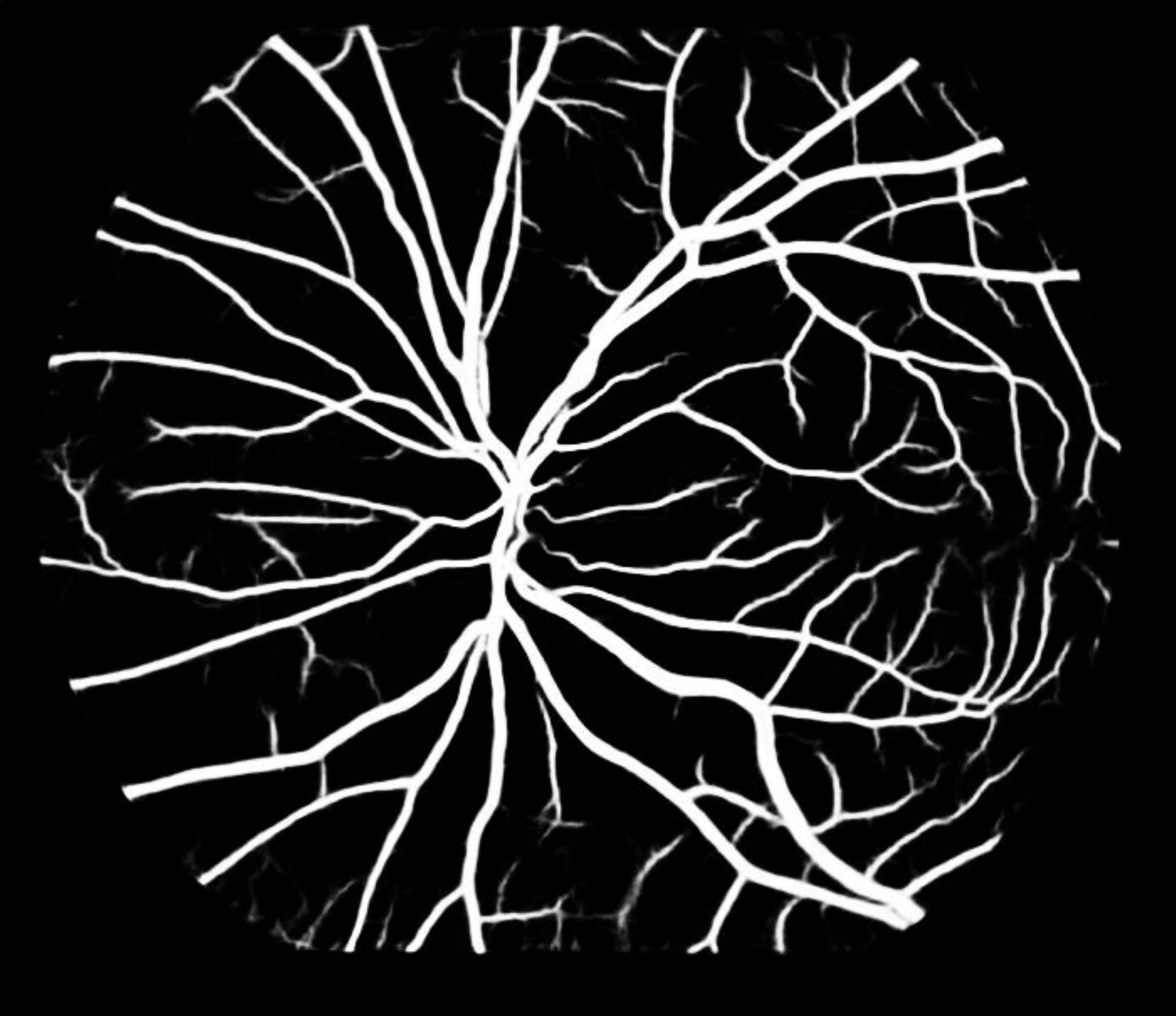}
}\subfigure{
\includegraphics[width=0.20\textwidth]{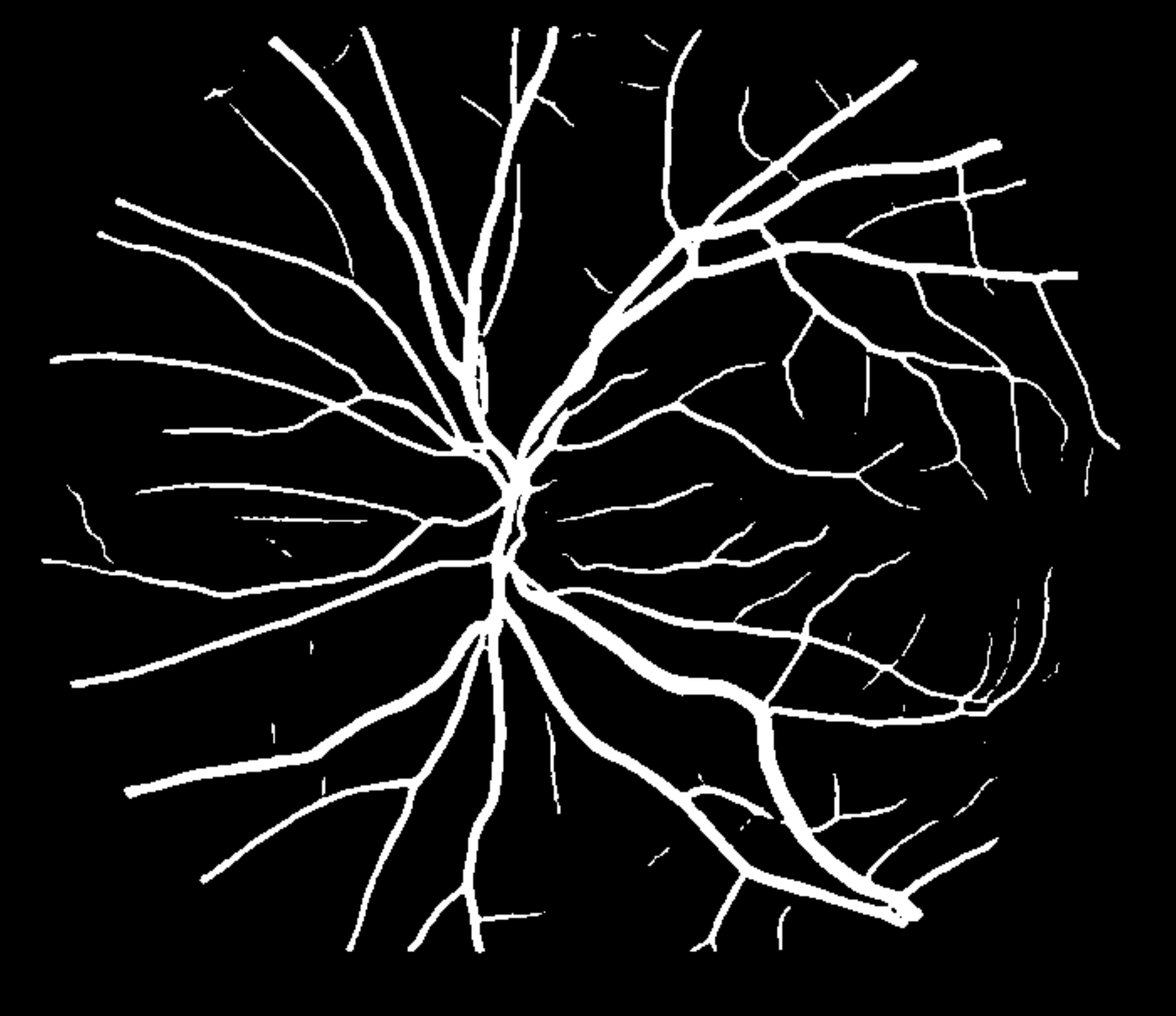}
}\\
\subfigure{
\includegraphics[width=0.20\textwidth]{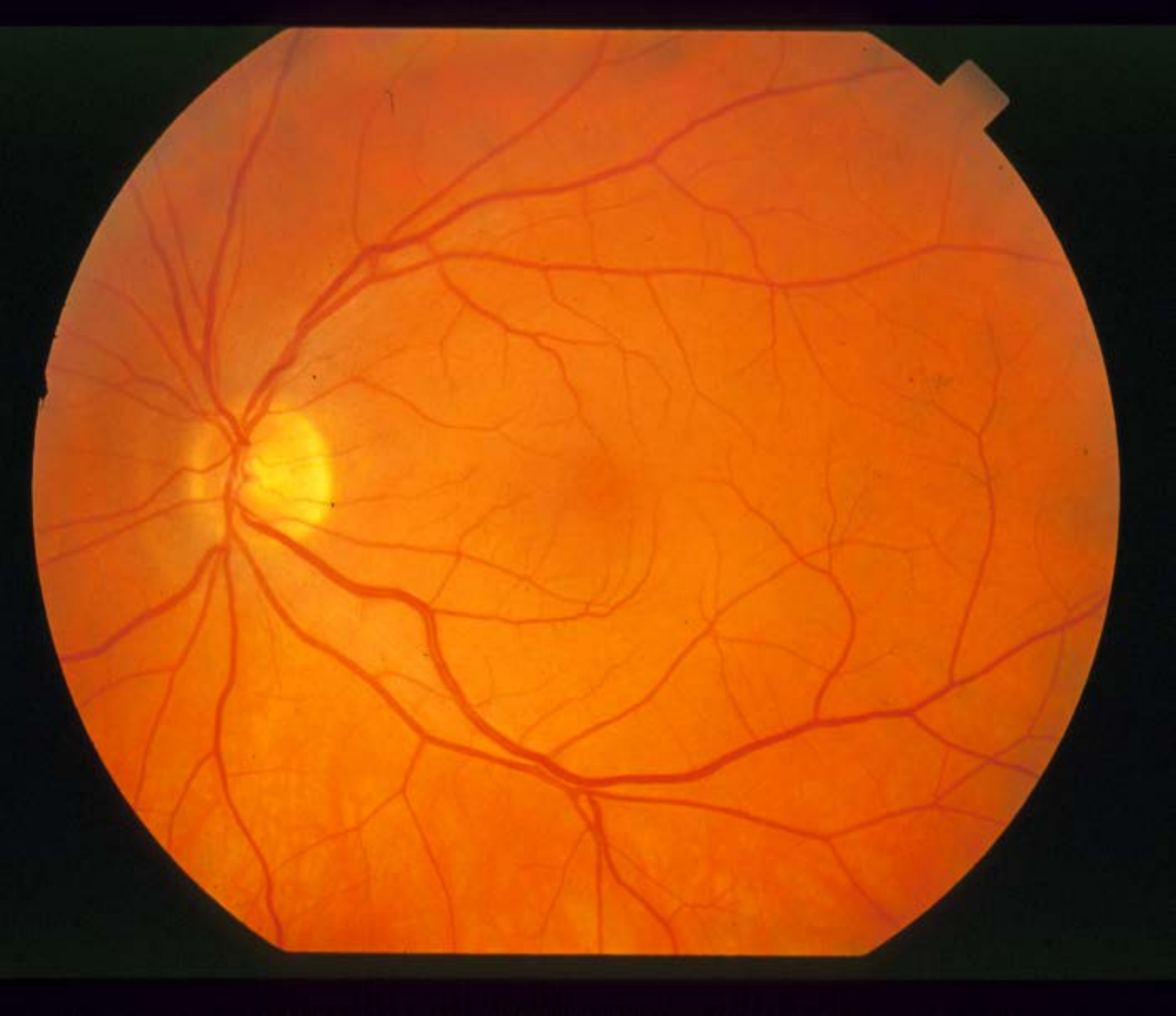}
}\subfigure {
\includegraphics[width=0.20\textwidth]{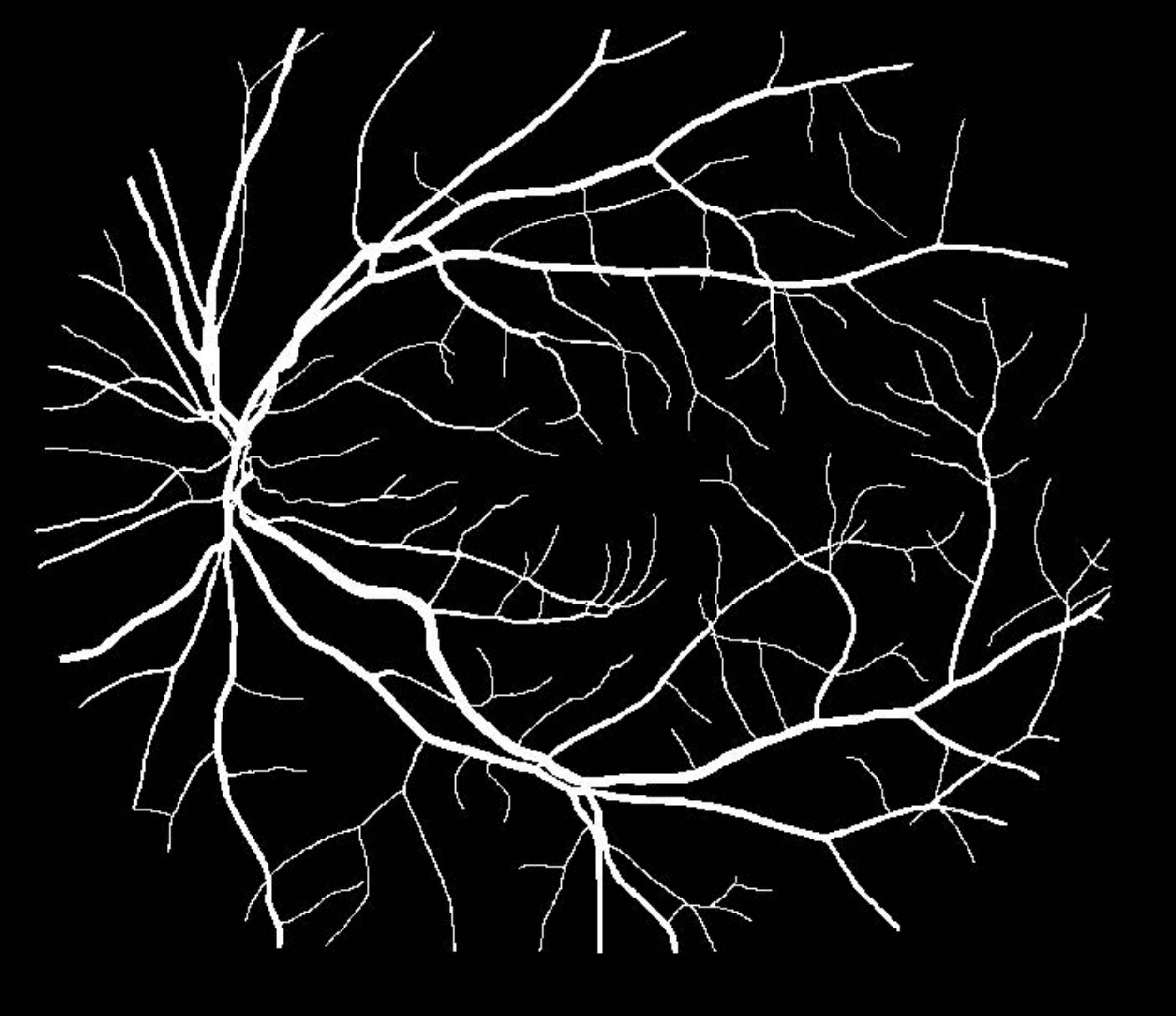}
}\subfigure{
\includegraphics[width=0.20\textwidth]{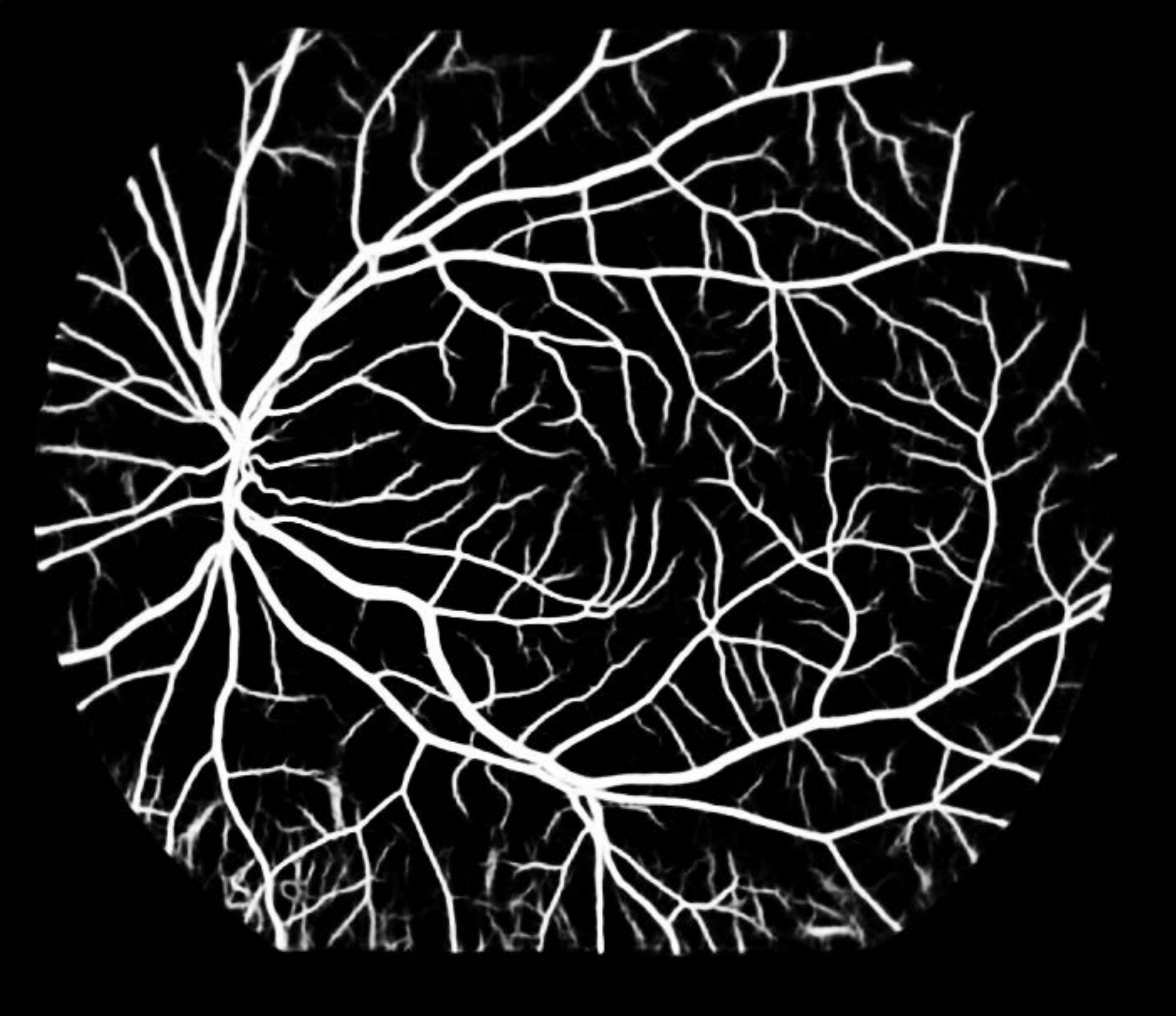}
}\subfigure{
\includegraphics[width=0.20\textwidth]{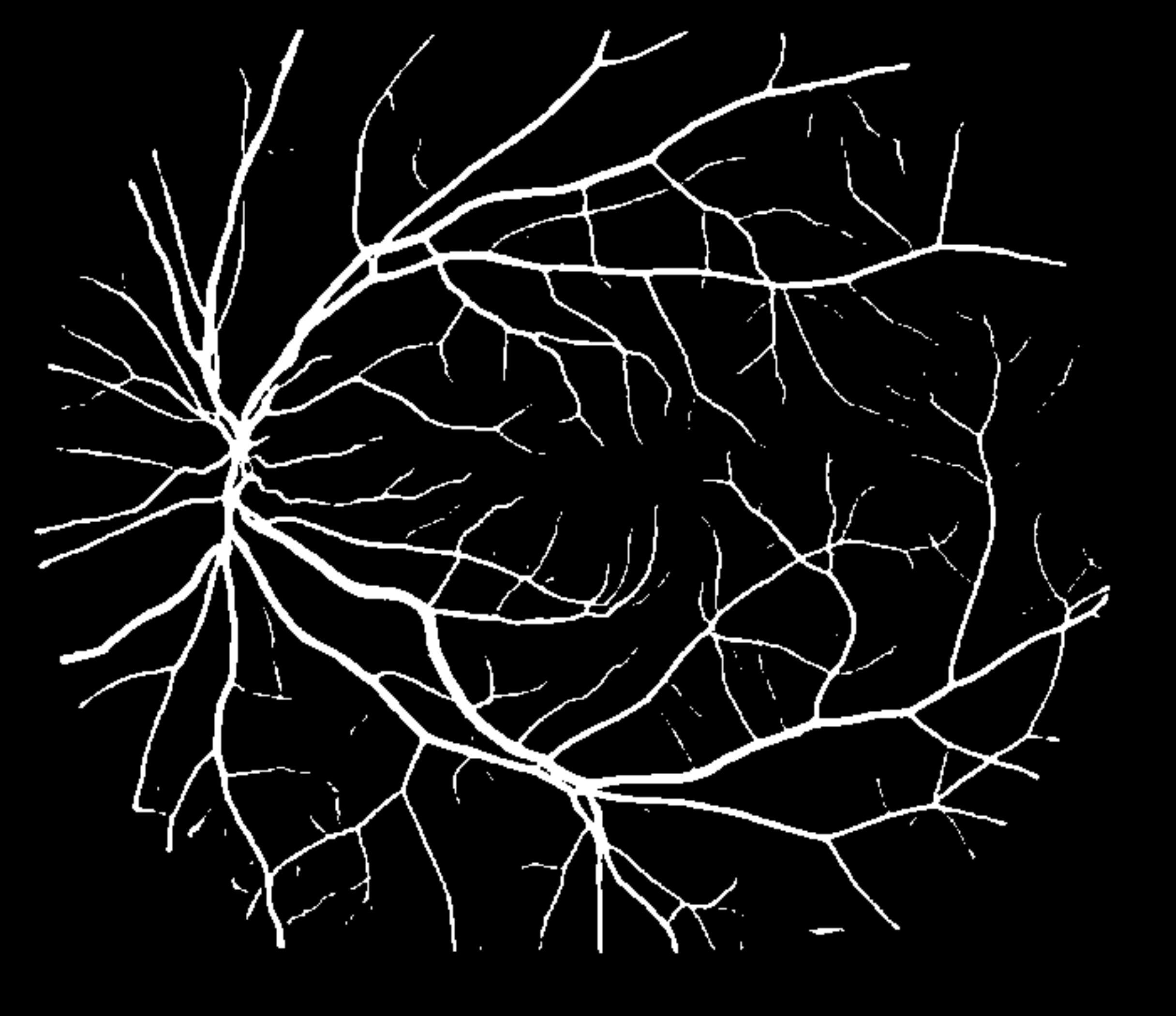}
}\\
\subfigure{
\includegraphics[width=0.20\textwidth]{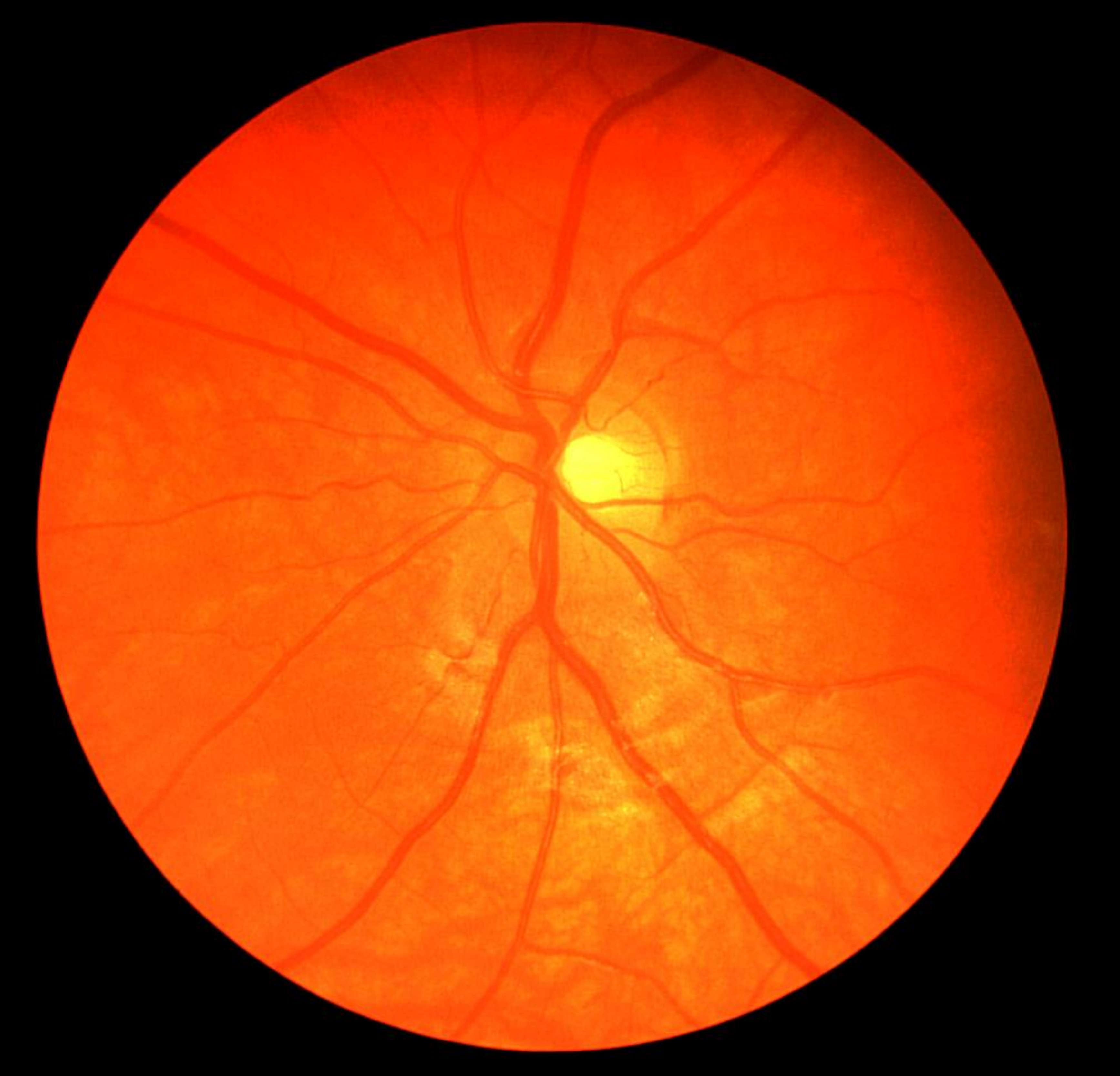}
}\subfigure {
\includegraphics[width=0.20\textwidth]{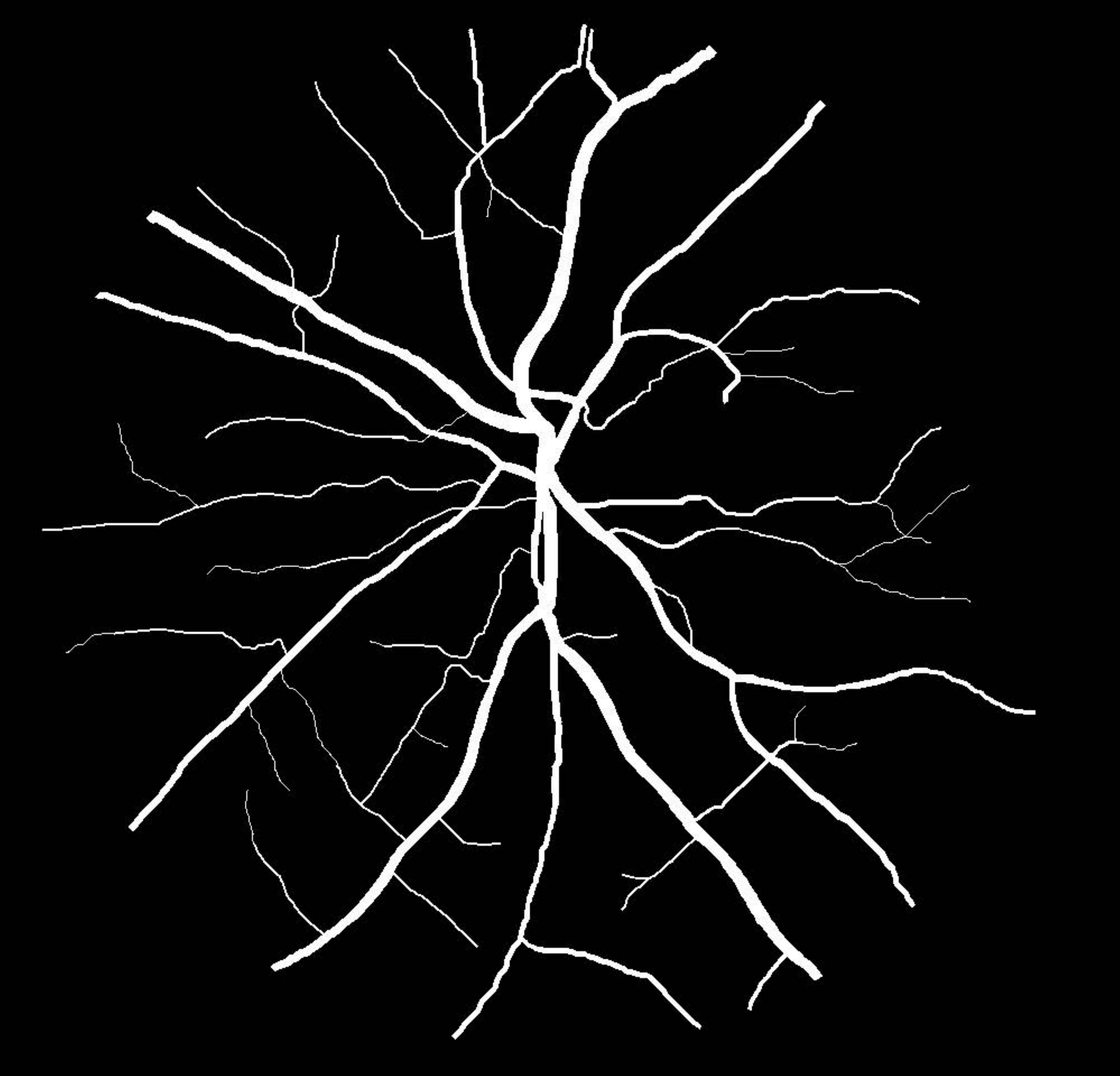}
}\subfigure{
\includegraphics[width=0.20\textwidth]{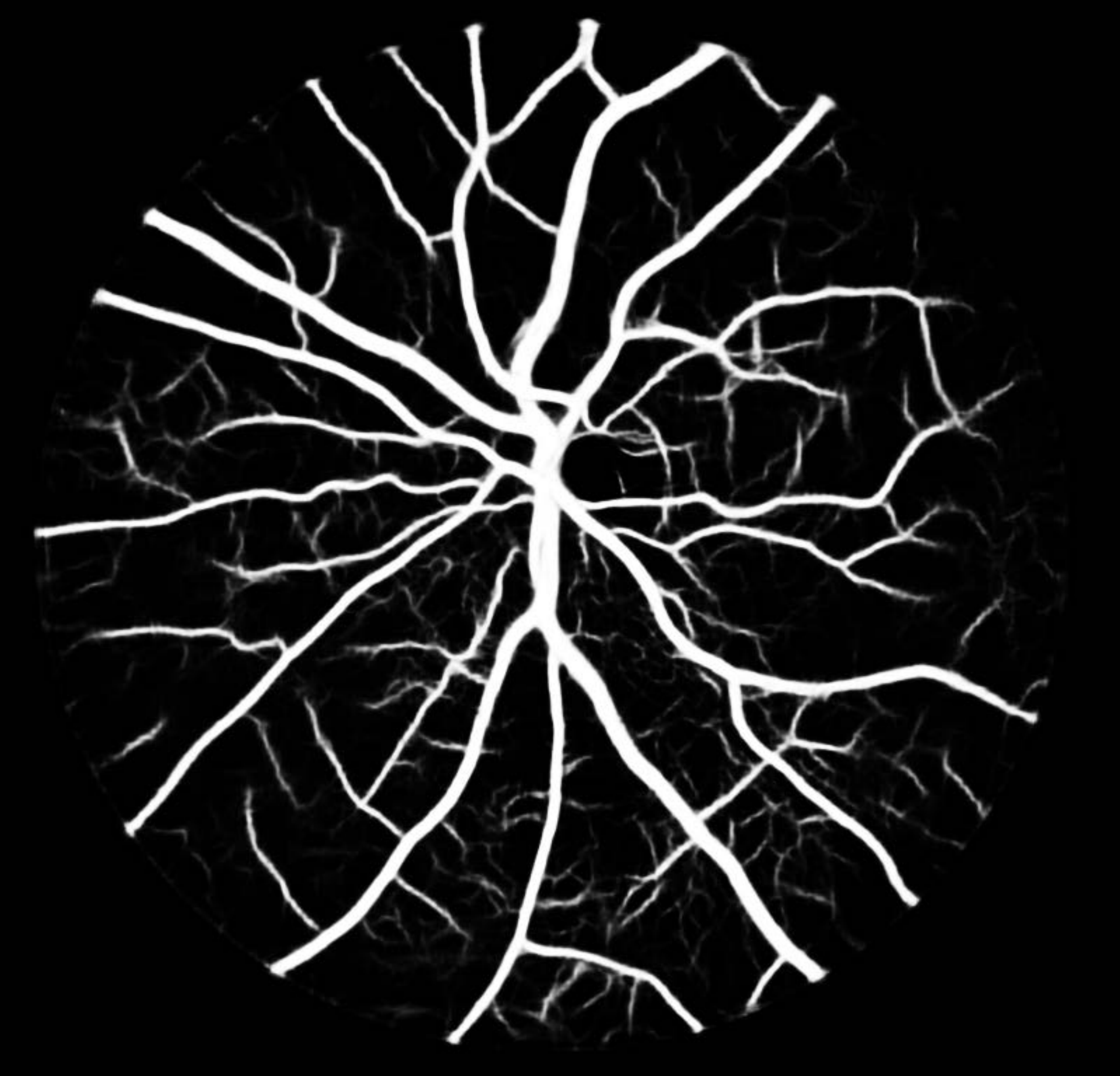}
}\subfigure{
\includegraphics[width=0.20\textwidth]{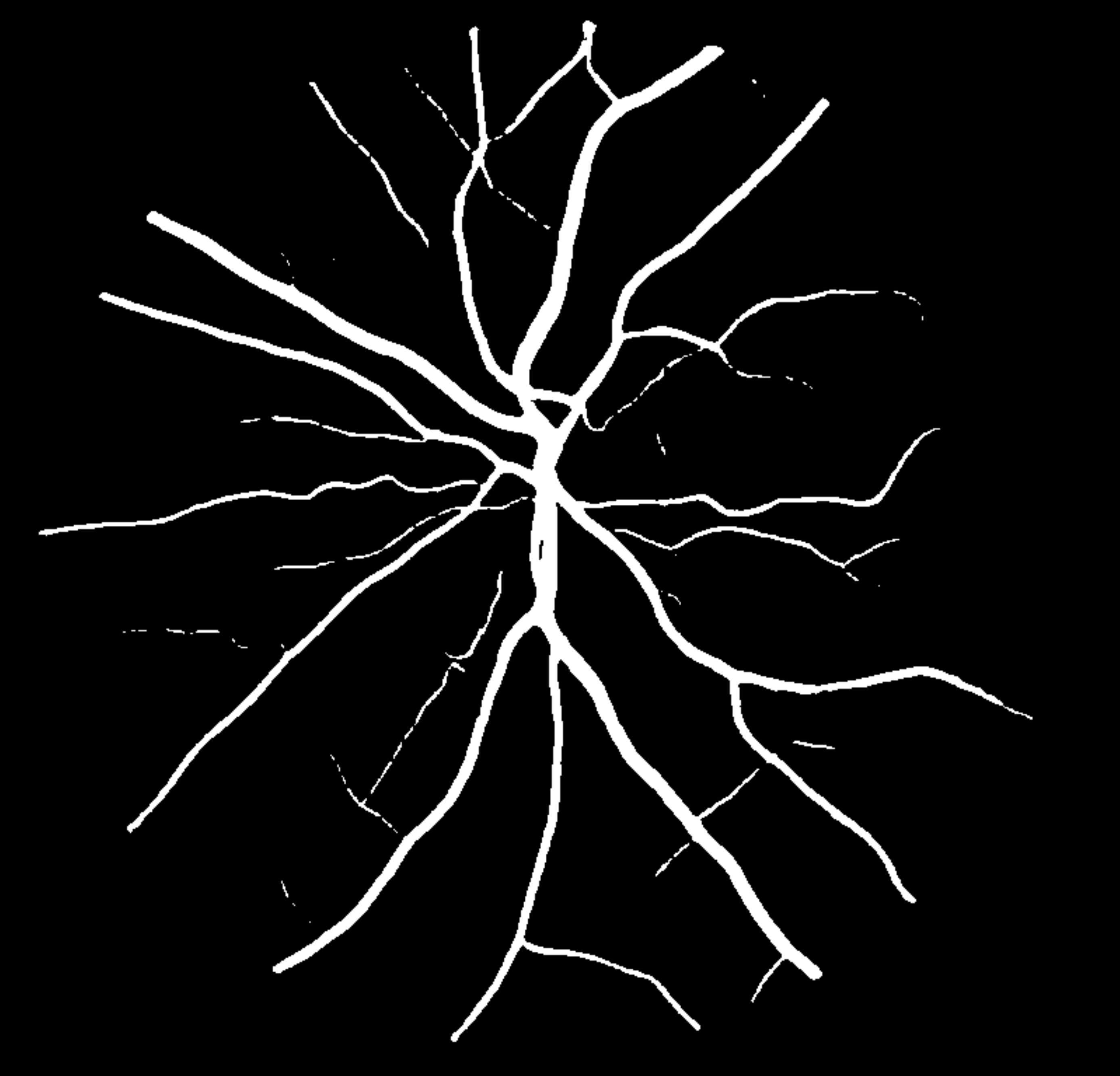}
}\\
\subfigure{
\includegraphics[width=0.20\textwidth]{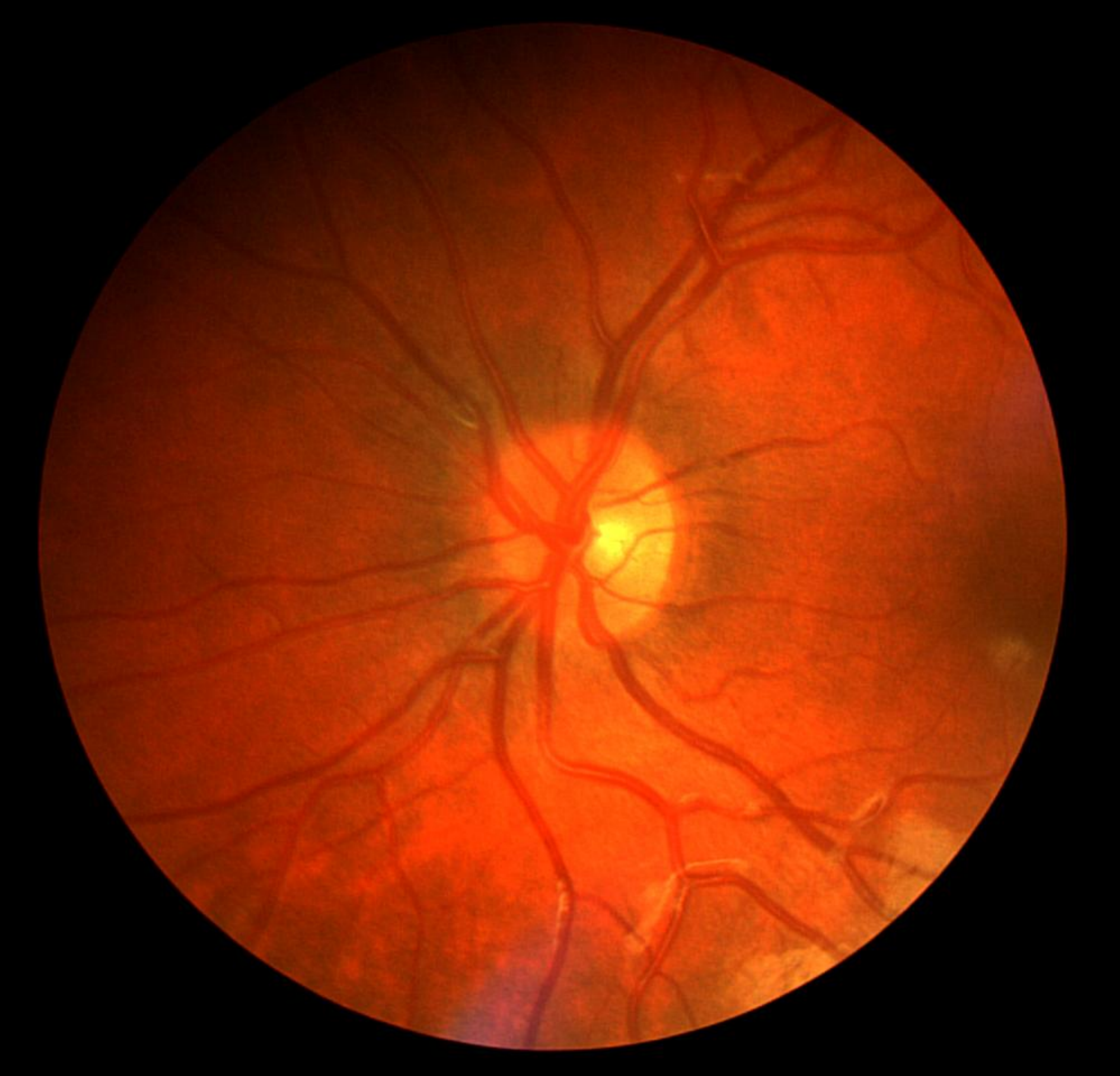}
}\subfigure {
\includegraphics[width=0.20\textwidth]{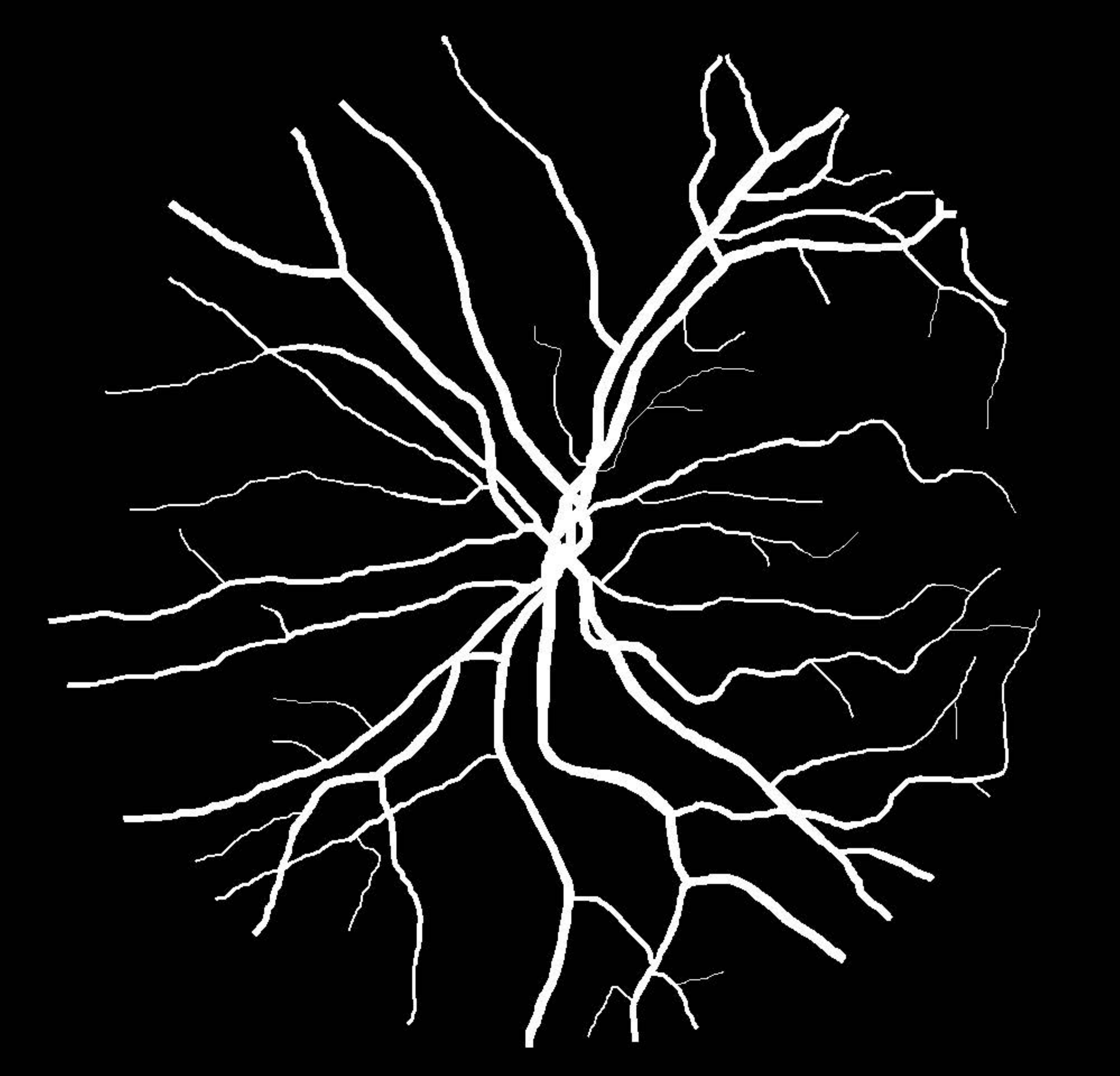}
}\subfigure{
\includegraphics[width=0.20\textwidth]{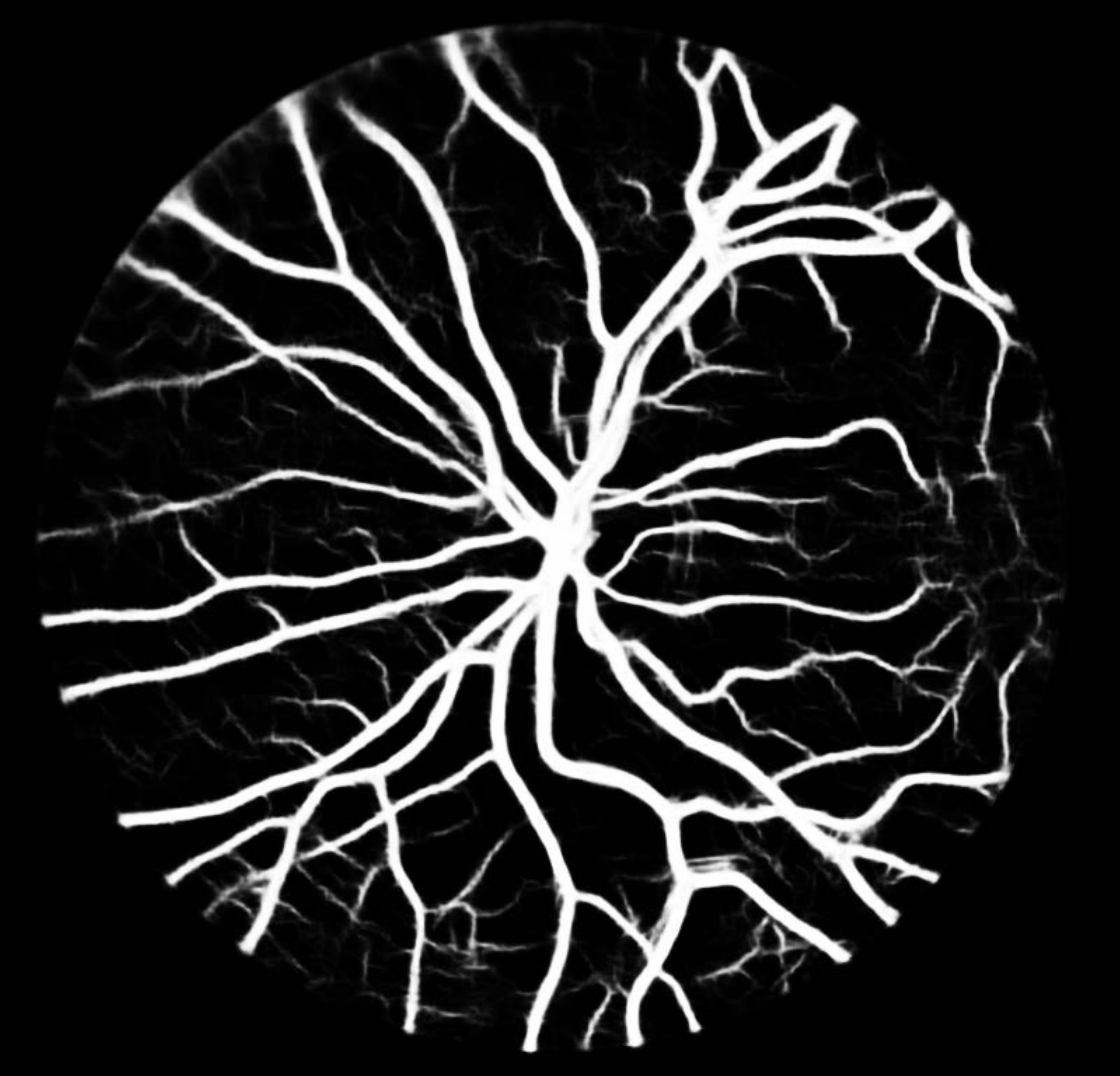}
}\subfigure{
\includegraphics[width=0.20\textwidth]{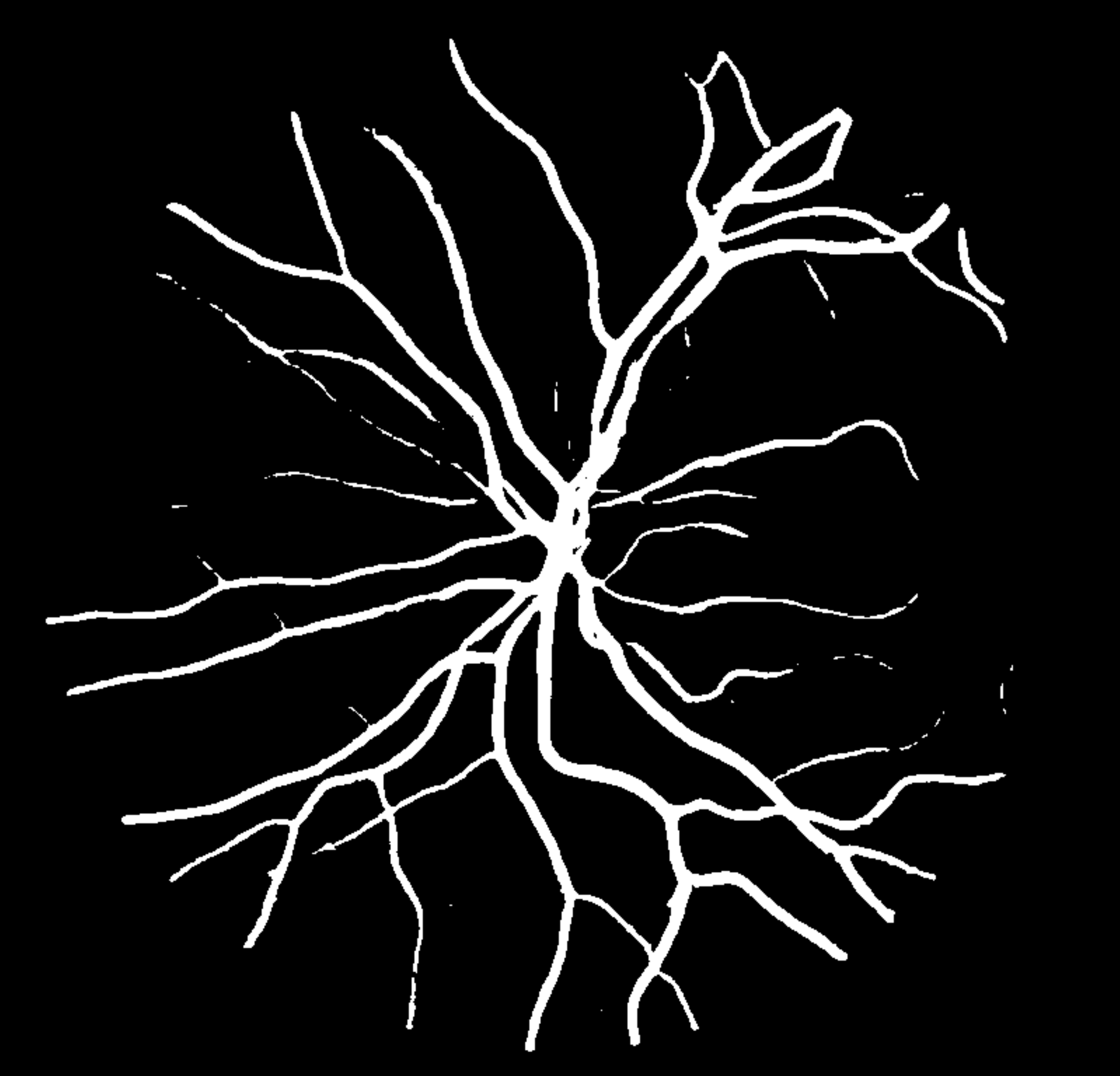}
}
\caption{(From left to right) retinal images, ground truth, segmentation probability map of BTS-DSN, and binary segmentation map of BTS-DSN. The top two images are from DRIVE dataset, the middle two images are from STARE dataset, and the bottom two images are from CHASE\_DB1 dataset.}
\label{fig:psdsn}
\end{figure}

The segmentation results of image-level input BTS-DSN on three datasets are shown in Figure~\ref{fig:psdsn}. We can observe from Figure~\ref{fig:psdsn} that the binary vessel segmentation results of BTS-DSN can recognize very coarse vessels. However, for thin vessels, the segmentation results are intermittent and many thin vessels cannot be identified.

\subsubsection{Comparison with the state-of-the-art}

In Tables~\ref{table:drive_result}, \ref{table:stare_result} and \ref{table:chasedb1_result}, we compare our approach with other state-of-the-art methods in terms of sensitivity, specificity, accuracy, AUC, MCC and F1-score on DRIVE, STARE and CHASE\_DB1 datasets.

We can observe from Table~\ref{table:drive_result} that our patch-level BTS-DSN achieved the best performance than other state-of-the-art methods in terms of MCC and F1-score on DRIVE dataset. Especially, our patch-level BTS-DSN model achieved an AUC score of 0.9806 which is higher than most methods. We can observe from Table~\ref{table:stare_result} that on STARE dataset, both our image-level and patch-level BTS-DSN achieved much higher SE, AUC, MCC, and F1-score over DRIU.
On CHASE\_DB1 dataset, our BTS-DSN achieved much higher scores in terms of SE, ACC and AUC over \cite{Li2016A} and \cite{jbhi2018}. Although MS-NFN achieves much higher SP and AUC, our BTS-DSN gets much higher AUC and SE.

\begin{table*}[!ht]
\begin{center}
\begin{threeparttable}
\caption{Performance comparison with state-of-the-art methods on the DRIVE dataset (best results shown in bold)}
\label{table:drive_result}
\begin{tabular}{llllllll}
\hline\noalign{\smallskip}
Methods & Year &  SE & SP & ACC & AUC & MCC &F1-score \\
\noalign{\smallskip}
\hline
\noalign{\smallskip}
2nd human expert & - & 0.7796 & 0.9717 & 0.9470 & N.A &0.7591 &0.7910\\
Wavelets\cite{soares2006retinal} & 2006 & 0.7104 & 0.9740 & 0.9404 & 0.9436 &N.A &0.7620 \\
Line Detector\cite{ricci2007retinal} & 2007 & 0.4966 & 0.9831 & 0.9212 & 0.8655 &N.A &0.6920 \\
SE\cite{dollar2013structured} & 2013 & 0.5309 & 0.9780 & 0.9211 & 0.8935 &N.A &0.6580\\
Kernel Boost\cite{becker2013supervised} & 2013 & 0.7133 & 0.9810 & 0.9469 & 0.9307 &N.A &0.8000\\
Patch Neural Network\cite{Li2016A} & 2015 & 0.7569 & 0.9816 & 0.9527 & 0.9738 & N.A & N.A \\
Active Contour Model\cite{7055281} &2015 &0.7420 &0.9820 &0.9540 &0.8620 &N.A &0.7820\\
DeepVessel\cite{Fu2016DeepVessel} & 2016 & 0.7603 & N.A & 0.9523 & N.A & N.A & N.A \\
Pixel CNN\cite{Liskowski2016Segmenting} & 2016 & 0.7811 & 0.9807 & 0.9535 & 0.9790 & N.A & N.A \\
DRIU\cite{driu} & 2016 & 0.7855 & 0.9799 & 0.9552 & 0.9793 & 0.7914 &0.8220 \\
Patch FCN\cite{Oliveira2017Augmenting} & 2017 & 0.7810 & 0.9800 & 0.9543 & 0.9768 & N.A & N.A\\
Patch FCN\cite{feng2017patch} & 2017 & 0.7811 & \bf{0.9839} &0.9560 & 0.9792 & N.A &0.8183\\
Three-stage FCN\cite{jbhi2018} &2018 &0.7631 &0.9820 &0.9538 &0.9750 &N.A &N.A\\
Modified U-net\cite{zhang2018} &2018 &\bf0.8723 &0.9618 &0.9504 &0.9799 &N.A &N.A\\
MS-NFN\cite{wu2018}        & 2018   &0.7844 &0.9819 &0.9567 &0.9807 &N.A &N.A\\
Patch FCN\cite{OLIVEIRA2018229}&2018 &0.8039 &0.9804 &\bf0.9576 &\bf0.9821&N.A &N.A\\
\hline
Image BTS-DSN  & 2018 &0.7800 & 0.9806 & 0.9551 & 0.9796 &0.7923 &0.8208\\
Patch BTS-DSN & 2018 &0.7891 & 0.9804 &0.9561 &0.9806 &\bf0.7964 &\bf0.8249\\
\hline
\end{tabular}
\begin{tablenotes}
\item[1] N.A = Not Available
\end{tablenotes}
\end{threeparttable}
\end{center}
\end{table*}

\begin{table*}[!ht]
\begin{center}
\begin{threeparttable}
\caption{Performance comparison with state-of-the-art methods on the STARE dataset (best results shown in bold)}
\label{table:stare_result}
\begin{tabular}{llllllll}
\hline\noalign{\smallskip}
Methods & Year &  SE & SP & ACC & AUC &MCC &F1-score \\
\noalign{\smallskip}
\hline
\noalign{\smallskip}
2nd human expert & - &\bf0.8952 &0.9384 &0.9349 & - &0.7454 &0.7600 \\
Line Detector\cite{ricci2007retinal} & 2007 &0.6307 &\bf0.9851 &0.9484 &0.9443 &N.A &0.7430\\
DeepVessel\cite{Fu2016DeepVessel} & 2016 & 0.7412 & N.A & 0.9585 & N.A & N.A & N.A \\
DRIU\cite{driu} & 2016 & 0.8036 &0.9845 & 0.9658 & 0.9773 &0.8110 &0.8310 \\
\hline
Image BTS-DSN  & 2018 &0.8201 &0.9828 &0.9660 &\bf0.9872 &0.8142 &0.8362\\
Patch BTS-DSN & 2018 &0.8212 &0.9843 & \bf0.9674 & 0.9859 & \bf0.8221 &\bf0.8421\\
\hline
\end{tabular}
\begin{tablenotes}
\item[1] N.A = Not Available
\end{tablenotes}
\end{threeparttable}
\end{center}
\end{table*}

\begin{table*}[!ht]
\begin{center}
\begin{threeparttable}
\caption{Performance comparison with state-of-the-art methods on the CHASE\_DB1 dataset (best results shown in bold)}
\label{table:chasedb1_result}
\begin{tabular}{llllllll}
\hline\noalign{\smallskip}
Methods & Year &  SE & SP & ACC & AUC &MCC &F1-score \\
\noalign{\smallskip}
\hline
\noalign{\smallskip}
2nd human expert  & - &\bf0.8315 &0.9745 &0.9615 & - &\bf0.7766 & 0.7970 \\
Ensemble Decision Trees\cite{Muhammad2012An}                             & 2012 & 0.7224 &0.9711 &0.9469 &0.9712 &N.A &N.A\\
COSFIRE\cite{Azzopardi2015Trainable} & 2015 & 0.7585 &0.9587 &0.9387 &0.9487 &N.A &N.A\\
Patch Neural Network\cite{Li2016A} & 2015 & 0.7507 & 0.9793 & 0.9581 & 0.9716 & N.A & N.A \\
Three-stage FCN\cite{jbhi2018} &2018 &0.7641 &0.9806 &0.9607 &0.9776 &N.A &N.A\\
MS-NFN \cite{wu2018}         &2018 &0.7538  &\bf0.9847  &\bf0.9637  &0.9825 &N.A &N.A\\
\hline
Image BTS-DSN &2018  &0.7888 &0.9801  &0.9627 &\bf0.9840 &0.7733 &\bf0.7983\\
\hline
\end{tabular}
\begin{tablenotes}
\item[1] N.A = Not Available
\end{tablenotes}
\end{threeparttable}
\end{center}
\end{table*}

\subsubsection{Cross-training}
We took cross-training experiments, i.e., training on one dataset and testing on another dataset, to show that our method can generalize well to other fundus images obtained with different cameras. The statistical results are shown in Table~\ref{table:cross_train}. When we train BTS-DSN on STARE dataset and test our model on DRIVE dataset, our method achieves the highest SE, ACC and AUC over \cite{Muhammad2012An} and \cite{Li2016A}. In addition, when we use STARE as test set and train BTS-DSN on other two datasets, our model achieves the highest SE, ACC and AUC over other two methods. Cross-training experiments show that our BTS-DSN can generalize well to real world application.

\begin{table*}[!htbp]
\begin{center}
\caption{Performance measure of cross-training (best results shown in bold)}
\begin{tabular}{cccccc}
\hline
Dataset &Segmentation &SE &SP &ACC &AUC\\
\hline
\multirow{2}{*}{DRIVE}
 &Trained on STARE\cite{Muhammad2012An} &0.7242 &0.9792 &0.9456 &0.9697\\
 &Trained on STARE\cite{Li2016A} &0.7273 &0.9810 &0.9486 &0.9677\\
 &Trained on STARE(Proposed method) &\bf0.7446 &0.9784 &\bf0.9502 &\bf0.9709 \\
 &Trained on CHASE\_DB1\cite{Li2016A}    &0.7307 &\bf0.9811 &0.9484 &0.9605\\
 &Trained on CHASE\_DB1(Proposed method)      &0.6960 &0.9699 &0.9377 &0.9523 \\
\hline
\multirow{2}{*}{STARE}
 &Trained on DRIVE\cite{Muhammad2012An} &0.7010 &0.9770 &0.9495 &0.9660\\
 &Trained on DRIVE\cite{Li2016A}        &0.7027 &0.9828 &0.9545 &0.9671\\
 &Trained on DRIVE(Proposed method)     &\bf0.7188 &0.9816 &\bf0.9548 &\bf0.9686\\
 &Trained on CHASE\_DB1\cite{Li2016A}   &0.6944 &\bf0.9831 &0.9536 &0.9620\\
 &Trained on CHASE(Proposed method)     &0.6799 &0.9808 &0.9501 &0.9517\\
\hline
\multirow{2}{*}{CHASE\_DB1}
 &Trained on STARE\cite{Muhammad2012An} &0.7103 &0.9665 &0.9415 &0.9565\\
 &Trained on STARE\cite{Li2016A}        &\bf0.7240 &0.9768 &0.9417 &0.9553\\
 &Trained on STARE(Proposed method)     &0.6726 &0.9710 &0.9411 &0.9511\\
 &Trained on DRIVE\cite{Li2016A}        &0.7118 &\bf0.9791 &0.9429 &\bf0.9628\\
 &Trained on DRIVE(Proposed method)     &0.6980 &0.9715 &\bf0.9441 &0.9565\\
\hline
\end{tabular}
\label{table:cross_train}
\end{center}
\end{table*}

\section{Conclusion}
In this paper, we propose a deeply-supervised fully convolutional neural network with bottom-top and top-bottom short connections, called BTS-DSN, for retinal vessel segmentation. We used short connections to alleviate the semantic gap between side-output layers, and experimental results have demonstrated effectiveness of the short connections when the backbone is VGGNet and ResNet-101. We have proved that the proposed BTS-DSN method can produce state-of-the-art results on DRIVE, STARE and CHASE\_DB1 datasets. However, the segmentation of the micro-vessel with one to several pixels wide is still a challenge and needs further exploration.

\section{Conflict of interest}
The authors declare that they have no conflict of interest.

\section{Acknowledgements}
This work is partially supported by the National Natural Science Foundation (61872200),  the National Key Research and Development Program of China (2016YFC0400709, 2018YFB1003405), the Natural Science Foundation of Tianjin (18YFYZCG00060).

\section*{References}

\bibliography{bts_dsn}

\end{document}